\documentclass[sigconf, screen]{acmart}

\AtBeginDocument{%
  }

\setcopyright{acmlicensed}
\copyrightyear{2025}
\acmYear{2025}
\acmDOI{XXXXXXX.XXXXXXX}
\acmConference[MM '25]{the 33rd ACM International Conference on Multimedia}{October 27--31,
  2025}{Dublin, Ireland}
\acmISBN{978-1-4503-XXXX-X/2018/06}

\usepackage{amsmath}

\begin{document}

\title{Tora2: Motion and Appearance Customized Diffusion Transformer for Multi-Entity Video Generation}


\author{Zhenghao Zhang}
\email{zhangzhenghao.zzh@alibaba-inc.com}
\affiliation{%
  \institution{Alibaba Group}
    \country{China}
}

\author{Junchao Liao}
\email{liaojunchao.ljc@alibaba-inc.com}
\orcid{0000-0003-4282-0843}
\affiliation{%
  \institution{Alibaba Group}
    \country{China}
}

\author{Xiangyu Meng}
\email{hzmengxiangyu@gmail.com}
\orcid{0009-0001-6224-7979}
\affiliation{%
  \institution{Alibaba Group}
    \country{China}
}

\author{Long Qin}
\email{ql362507@alibaba-inc.com}
\affiliation{%
  \institution{Alibaba Group}
    \country{China}
}

\author{Weizhi Wang}
\email{wangweizhi.wwz@alibaba-inc.com}
\affiliation{%
  \institution{Alibaba Group}
    \country{China}
}


\begin{abstract}
Recent advances in diffusion transformer models for motion-guided video generation, such as Tora, have shown significant progress. In this paper, we present Tora2, an enhanced version of Tora, which introduces several design improvements to expand its capabilities in both appearance and motion customization. Specifically, we introduce a decoupled personalization extractor that generates comprehensive personalization embeddings for multiple open-set entities, better preserving fine-grained visual details compared to previous methods. Building on this, we design a gated self-attention mechanism to integrate trajectory, textual description, and visual information for each entity. This innovation significantly reduces misalignment in multimodal conditioning during training. Moreover, we introduce a contrastive loss that jointly optimizes trajectory dynamics and entity consistency through explicit mapping between motion and personalization embeddings. Tora2 is, to our best knowledge, the first method to achieve simultaneous multi-entity customization of appearance and motion for video generation. Experimental results demonstrate that Tora2 achieves competitive performance with state-of-the-art customization methods while providing advanced motion control capabilities, which marks a critical advancement in multi-condition video generation.
 
\end{abstract}



\begin{teaserfigure}
  \centering
  \includegraphics[width=0.95\textwidth]{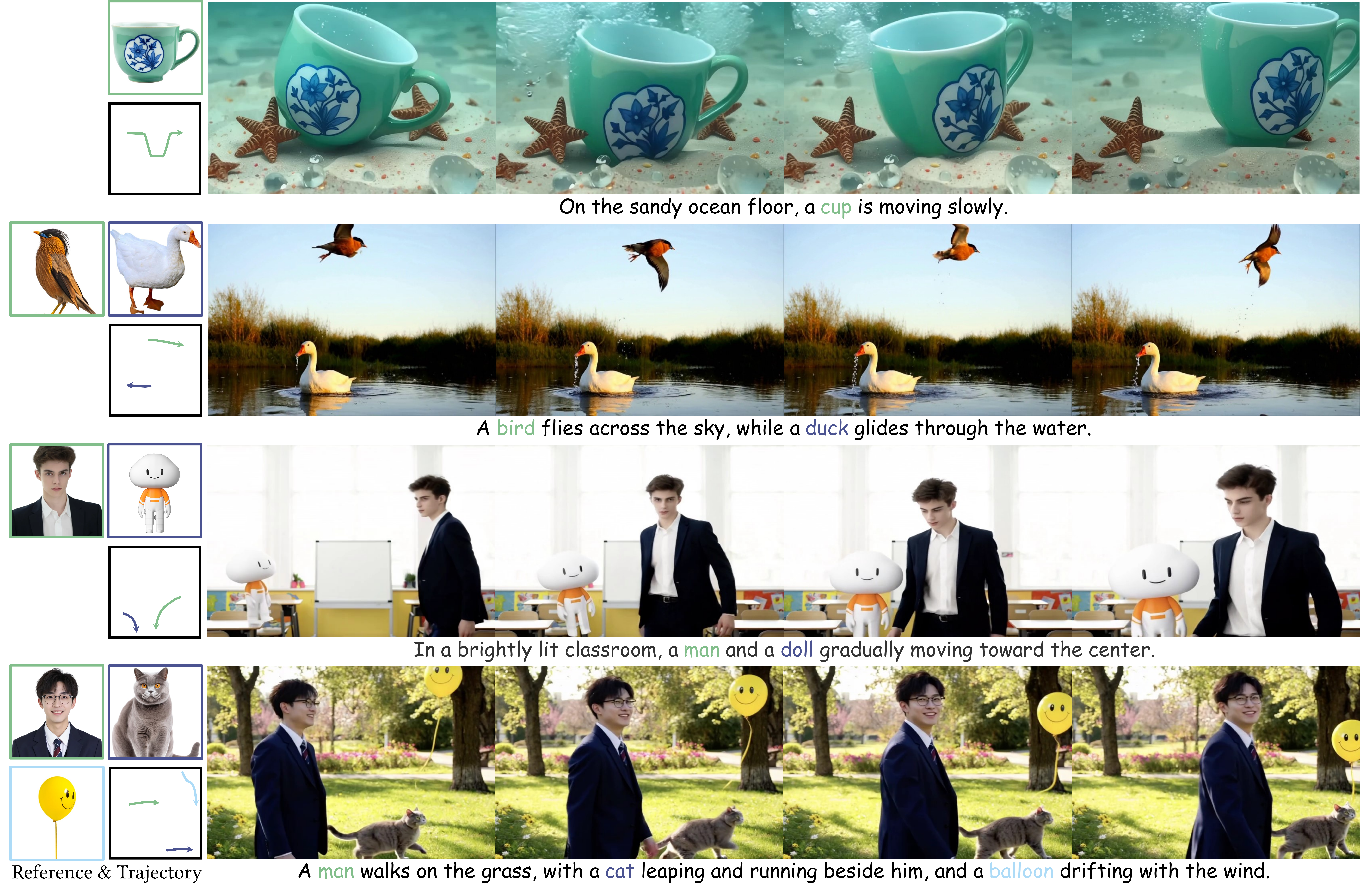}
  \caption{Given text prompts, motion trajectories, and reference images for  human entities (e.g., man, woman) and non-human entities (e.g., cup, cat, duck), Tora2 generates natural motions following the specified trajectories and textual descriptions for each entity while preserving their unique identities. We highly recommend viewing the videos in the appendix.}
  \Description{teaser figure}
  \label{fig:teaser}
  \vspace{3mm}
\end{teaserfigure}


\maketitle

\section{Introduction}

Recent advancements in large-scale video Diffusion Transformer (DiT) models~\cite{sora2024,polyak2024movie} have notably transformed the landscape of text-to-video generation. This innovation facilitates a broad array of downstream controllable scenes, making a significant impact on the customization of entity appearance or motion pattern. For instance, Tora~\cite{zhang2024toratrajectoryorienteddiffusiontransformer} integrates trajectory encoder with motion fusion techniques to implement motion controls within DiT frameworks. Meanwhile, ConsisID~\cite{yuan2024identitypreservingtexttovideogenerationfrequency} utilizes a frequency-aware identity preservation DiT model tailored for videos of individual person. Additionally, Video Alchemist~\cite{chen2025multisubjectopensetpersonalizationvideo} combines multiple conditional reference images to facilitate the personalization of videos featuring multiple subjects in open-set contexts. Collectively, these innovations significantly enhance the practical applications of video generation across a variety of scenarios.

Recent research efforts~\cite{wei2023dreamvideocomposingdreamvideos, wei2024dreamvideo2zeroshotsubjectdrivenvideo,wu2024motionboothmotionawarecustomizedtexttovideo,jiang2023videoboothdiffusionbasedvideogeneration} have investigated the generation of customized video content tailored to both appearance and motion. These approaches are classified into tuning-based methods~\cite{wei2023dreamvideocomposingdreamvideos,wu2024motionboothmotionawarecustomizedtexttovideo} and tuning-free methods~\cite{wei2024dreamvideo2zeroshotsubjectdrivenvideo}. Tuning-based methods optimize model parameters and manipulate attention maps for appearance adaptation and motion dynamics, while tuning-free methods learn subject appearance and motion patterns during training, enabling integration without additional fine-tuning. Despite notable progress, challenges remain. Primarily, the reliance on U-Net architectures~\cite{ronneberger2015u} limits application to emerging DiT models, reducing generation capabilities. Furthermore, current methods focus on single-entity customization, complicating the simultaneous manipulation of multiple entities with diverse motion control.

To address these challenges, we propose Tora2, a tuning-free DiT method specifically designed to maintain the fidelity of multiple entities while adhering to predefined motion trajectories. Existing open-set personalization approaches, such as Video Alchemist, predominantly utilize low-frequency global features extracted by DINOv2~\cite{oquab2024dinov2learningrobustvisual} for appearance injection. However, due to the human eye's sensitivity to high-frequency details, reliance on low-frequency information often leads to insufficient content similarity, particularly in multi-entity scenarios.
To enhance this, we introduce a decoupled personalization extractor (DPE) that integrates both low- and high-frequency details to create comprehensive personalization embeddings, thereby preserving fine-grained visual information more effectively. Specifically, high-frequency features are first extracted using facial analysis networks~\cite{deng2019arcface} and ReID networks~\cite{liu2022opening}, enabling human-specific and object-specific customization. Our experiments demonstrate that this decoupled approach improves consistency in human face representations. These features are transformed by appearance adapters, combined with global features, and queried using a Q-Former network~\cite{li2022blipbootstrappinglanguageimagepretraining} to generate personalization embeddings. For multi-entity control, a gated self-attention mechanism is utilized to fuse the personalization embeddings with corresponding motion embeddings and textual entity tokens. This mechanism enables the integration of appearance, motion, and textual information, ensuring effective learning across multiple entities. Additionally, we introduce a contrastive learning framework to strengthen cross-modal alignment between the entities and the motion patterns extracted by Tora.

To realize concurrent customization of motion and appearance, we integrate motion embeddings and personalization embeddings at distinct stages within the diffusion transformer. Specifically, motion embeddings are infused through an adaptive layer normalization~\cite{huang2017arbitrary} mechanism that modulates video latents prior to the 3D full-attention computation. Meanwhile, personalization embeddings are introduced via an independent cross-attention module positioned post the primary attention layer. Through systematic exploration of multiple embedding injection sequences, we validate that this hierarchical injection strategy achieves optimal generative performance in terms of both motion fidelity and identity preservation.

In this work, we merge the data collection pipelines from Video Alchemist and Tora, creating a dataset of 1.1 million video clips featuring diverse entities with fluid motion patterns, specifically curated for training Tora2. Evaluation on the MSRVTT-Personalization benchmark~\cite{chen2025multisubjectopensetpersonalizationvideo} reveals that Tora2 excels in multi-entity video customization with precise trajectory control.

Our contributions can be summarized as follows:
\begin{itemize}
\item We propose Tora2, the first multi-entity customized video diffusion transformer framework. As shown in Figure~\ref{fig:teaser}, Tora2 allows for both appearance and motion trajectory control.

\item We introduce a decoupled personalization extractor designed to learn fine-grained personalization embeddings for open-set entities, enhancing both subject and face similarity in multi-entity generation.

\item We design a novel binding strategy and a contrastive learning approach to ensure consistent and aligned representation of entities, motion patterns, and textual descriptions.

\item Experimental evaluations show that Tora2 achieves competitive performance with state-of-the-art customization methods, while also introducing advanced motion trajectory control.

\end{itemize}
\section{Related work}
\subsection{Video Diffusion Model}
The evolution of video generation models ~\cite{DBLP:journals/corr/abs-2311-04145, wang2024tf,wang2023modelscope} has been driven by progressive architectural innovations. Early approaches like Stable Video diffusion~\cite{DBLP:journals/corr/abs-2311-15127}, VideoCraft~\cite{Chen2023VideoCrafter1OD}, and Animate Anything ~\cite{DBLP:journals/corr/abs-2311-12886} primarily used U-Net-based architectures with 3D convolutional layers for temporal modeling. However, these frameworks struggled to maintain motion coherence in extended sequences and had limitations in dynamic scene understanding. Recent advancements have shifted towards transformer-based architectures, as seen in models such as Sora~\cite{sora2024}, MovieGen~\cite{polyak2024movie}, Vidu~\cite{DBLP:journals/corr/abs-2405-04233}, and HunyuanVideo~\cite{kong2025hunyuanvideosystematicframeworklarge}. These models utilize a 3D causal VAE~\cite{yu2023magvit} to handle the encoding and decoding of raw video data, significantly increasing the compression ratio. Next, they utilize the DiT model architecture, renowned for its exceptional scalability, and integrate a 3D full-attention mechanism, substantially enhancing the potential for controllable video generation.

\begin{figure*}[!t]
    \centering
    \includegraphics[width=0.95\textwidth]{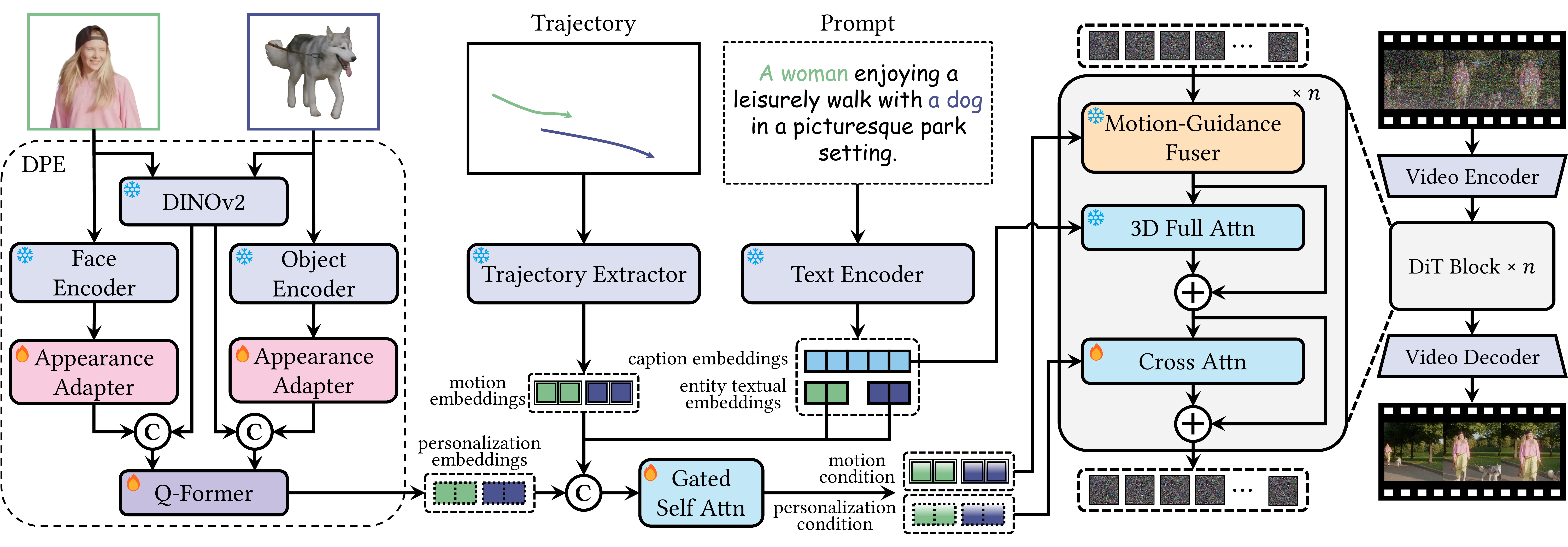}
    \caption{An overview of Tora2, which consists of a decoupled personalization extractor~(DPE), a trajectory extractor, a video diffusion transformer, and a gated self-attention mechanism for entity binding.  
The DPE generates personalization embeddings by combining high-frequency detail information extracted via a decoupled strategy for both human and non-human objects with the low-frequency semantic features obtained from DINOv2. The trajectory extractor encodes provided trajectories into motion embeddings, which are bound to visual entities using a gated self-attention mechanism. These bound motion and personalization conditions are then fed into the video diffusion transformer, employing a motion-guidance fuser and an additional cross-attention layer to achieve both motion and appearance customization for multi entities.
    }
    \label{fig:pipeline}   
\end{figure*}

\subsection{Trajectory Control in Video Generation}
Video generation through trajectory control ~\cite{zhao2022thin, mallya2022implicit, wang2023videocomposer,DBLP:journals/corr/abs-2311-12886} has become increasingly popular for its precise motion control capabilities. Early methods like DragNUWA~\cite{yin2023dragnuwa} and MotionCtrl~\cite{wang2023motionctrl} use flow maps and trajectory coordinates transformed into dense vector maps as guidance signals. TrailBlazer~\cite{DBLP:journals/corr/abs-2401-00896} utilizes bounding boxes to steer subject motion during video generation, while DragAnything~\cite{wu2024draganythingmotioncontrolusing} combines entity representation extraction with a segmentation model to enable entity-level control. LeViTor~\cite{wang2024levitor3dtrajectoryoriented} utilizes keypoint trajectory maps that are augmented with depth information to guide 3D motion, effectively capturing and representing object movements. Tora~\cite{zhang2024toratrajectoryorienteddiffusiontransformer} uses a motion variational autoencoder~\cite{yu2023magvit} to integrate trajectory vectors into DiT architectures, ensuring motion information is retained across frames. Despite advances in motion control, these methods can only manipulate the given reference frame, lacking the ability for appearance customization.

\subsection{Appearance Control in Video Generation}
Recent advancements have expanded model personalization techniques ~\cite{ruiz2023dreamboothfinetuningtexttoimage, wei2023dreamvideocomposingdreamvideos, wei2024dreamvideo2zeroshotsubjectdrivenvideo, huang2025conceptmastermulticonceptvideocustomization} for video synthesis. DreamVideo~\cite{wei2023dreamvideocomposingdreamvideos} utilizes a tuning-based approach that simultaneously personalizes both identity and motion. VideoBooth~\cite{jiang2023videoboothdiffusionbasedvideogeneration} leverages foreground segmentation via Grounded-SAM~\cite{ren2024groundedsamassemblingopenworld} to construct personalized embeddings through a coarse-to-fine strategy. StoryDiffusion~\cite{zhou2024storydiffusion} uses a stable self-attention mechanism alongside a semantic motion predictor to guarantee smooth transitions and maintain consistent identity throughout its process. ID-Animator~\cite{he2024idanimatorzeroshotidentitypreservinghuman} combines IP-Adapter~\cite{ye2023ipadaptertextcompatibleimage} with AnimateDiff~\cite{guo2023animatediff}, employing joint optimization for facial personalization. Meanwhile, ConsisID~\cite{yuan2024identitypreservingtexttovideogenerationfrequency} separates facial features into low-frequency global features and high-frequency intrinsic features, integrating them within a DiT-based framework to maintain identity integrity. Unlike these single-concept methodologies, Video Alchemist~\cite{chen2025multisubjectopensetpersonalizationvideo} achieves multi-subject personalization by employing cross-attention fusion of reference images and textual prompts within a Diffusion Transformer. ConceptMaster~\cite{huang2025conceptmastermulticonceptvideocustomization} integrates visual and textual embeddings through a Decoupled Attention Module, subsequently incorporating these embeddings into the diffusion model using a multi-concept injector. In our research, we enhance appearance control using a decoupled personalization extractor and implement a novel binding strategy, achieving advanced multi-entity video customization with trajectory control.

\section{Methodology}
Our goal is to achieve multi-entity controllable video generation with appearance and trajectory conditioning. In this section, we first outline the diffusion transformer model and Tora framework. Then, we introduce Tora2, explaining how personalization embeddings for open-set entities are obtained, linked to entity words and trajectories, and injected as conditions. Finally, we describe the construction of the training dataset.

\subsection{Preliminary}

\textbf{Diffusion Transformer~(DiT).}
The DiT~\cite{peebles2023scalable} represents a breakthrough by integrating the robustness of diffusion models with the advanced capabilities of transformer architectures~\cite{DBLP:conf/nips/VaswaniSPUJGKP17}. This novel architecture is designed to overcome the inherent limitations of conventional U-Net-based latent diffusion models (LDMs), thereby enhancing their performance, versatility, and scalability. While maintaining the overall structure consistent with existing LDMs, the significant innovation involves substituting the U-Net with a transformer architecture to learn the denoising objective function:
\begin{equation}
    \mathcal{L}_\epsilon = ||\epsilon - \epsilon_\theta(z_t, t, c)||^2_2,
    \label{eq:training_objective}
\end{equation}
Here, $\epsilon_\theta(\cdot)$ signifies the output of DiT. The condition $c$ is guided into the DiT blocks using cross-attention for adjustment and $z_t$ denotes the noisy hidden state.
 
\noindent \textbf{Tora.}
Tora is the first trajectory-oriented Diffusion Transformer framework that unifies textual, visual, and trajectory conditions for scalable video generation with precise motion guidance. Its core innovation lies in encoding arbitrary trajectories into motion patches aligned with DiT's input tokens. Given a trajectory $traj=\left \{ (x_i, y_i) \right \} _{i=0}^{L-1} $ where $(x_i, y_i)$ denotes spatial coordinates at frame $i$, Tora employs a 3D motion VAE to extract compact motion embeddings $m \in \mathbb{R}^{l \times h \times w \times c}$ matching video latent dimensions, where $l$, $h$, $w$, $c$ denote compressed video length, height, width, and channels. A motion-guidance fuser then dynamically injects the motion embeddings into DiT blocks, which can be formulated as:

\begin{equation}
    z_{t} = \gamma \cdot z_{t} + \beta,
    \label{eq:training_objective}
\end{equation}

where the scale $\gamma$ and shift $\beta$ are converted from $m$ using two convolution layers that are initialized to zero. By training on a carefully curated large-scale video dataset, Tora achieves state-of-the-art performance, significantly outperforming existing motion-controllable methods.

\subsection{Tora2}

As shown in Figure~\ref{fig:pipeline}, Tora2 is a latent Diffusion Transformer that integrates extraction and injection processes for entity and trajectory inputs. Retaining Tora's trajectory conditioning architecture, we emphasize its novel personalized representation extraction method and binding strategy, which associates entities with corresponding text and trajectory conditions.

\noindent \textbf{Open-set personalization embeddings extraction.} To ensure the model can process multiple entities with high fidelity, deriving accurate visual representations from concept images $\left \{ I_{i} | i=1\dots N  \right \} $ is crucial. Previous research often employs the output from the last layer of the CLIP~\cite{radford2021learning} or DINOv2 image encoder as personalization embeddings $\left \{ p_{i}|p_{i} = \varepsilon _{\mathrm{img}}(I_{i}),  i=1\dots N \right \} $. Although these features encapsulate robust semantic information, they lack sufficient intrinsic identity representation, which often leads to unsatisfactory fidelity in multi-entity generation. Additionally, these features usually yield inadequate alignment with the representation space of diffusion models.

To address these limitations, we propose a decoupled personalization extractor. More specifically, we first utilize facial recognition and ReID backbones to extract identity-strong features for human-specific and generic objects, respectively. Decoupling is justified by the sensitivity of our eyes to facial details, with experiments demonstrating that this method enhances facial similarity in our scenario. Two additional appearance adapters are then employed to project these high-frequency features into a unified feature space. These identity-focused features are then concatenated with the features captured by the DINOv2 image encoder, which are semantically robust, thereby creating a comprehensive visual representation:
\begin{equation}
\left \{ v_{i}|v_{i} = \mathrm{Concat}(~f_{\theta } (\varepsilon _{\mathrm{prior}}(I_{i})), ~~\varepsilon _{\mathrm{img}}(I_{i})~) \right \}_{i=1}^{N}, 
\end{equation}
where $\varepsilon _{prior}$ denotes the corresponding pretrained class-specific recognition encoder and $f_{\theta }$ represents the appearance adapter. To facilitate better alignment with the diffusion transformer, we integrate a learnable Q-Former architecture, which consists of stacked cross-attention layers and feedforward networks. The comprehensive visual representation is used as a key-value corpus, and the Q-Former is leveraged to generate personalization embeddings for each concept:
\begin{equation}
\left \{ p_{i}|p_{i} = Q(v_{i}),  i=1\dots N  \right \} ,
\end{equation}
This approach significantly enhances the model's ability to maintain high fidelity when processing multiple concepts simultaneously.

\noindent \textbf{Binding of entity with the word and trajectory.} Empirical evidence~\cite{chen2025multisubjectopensetpersonalizationvideo} has shown that unbinding visual concepts from their corresponding textual descriptions leads to misalignment in multi-entity personalization. Therefore, linear projection is used to extend the personalization embeddings with textual information. Our framework extends this requirement to trajectory-conditioned generation, necessitating precise trimodal alignment between entity, text, and motion trajectory. We address this through a gated self-attention that ensures spatially consistent allocation of customized entities along designated trajectories.

For a given personalization embedding $p_{i}\in  R^{l_{p}\times d } $, it is associated with a trajectory characterized by motion embedding $m_{i}\in  R^{l_{m}\times d } $ and word tokens $c_{i}\in  R^{l_{c}\times d } $. In this context, $l_{p}$, $l_{m}$, $l_{c}$, and $d$ denote the number of tokens per reference image, trajectory tokens, text tokens, and token dimensions, respectively. We initiate the process by concatenating these three modal tokens and subsequently employ a self-attention mechanism integrated with two gated mechanisms to facilitate cross-modal interaction:
\begin{align}
\hat{p_{i}},  \hat{m_{i}}, \hat{c_{i}} &= \mathrm{SelfAttention}(\mathrm{Concat}(p_{i}, m_{i}, c_{i})), \nonumber \\
p_{i} &= p_{i} + G_{p}(\hat{p_{i}}), \nonumber \\ 
m_{i} &= m_{i} + G_{m}(\hat{m_{i}}), \tag{5}
\end{align}
where $G_{m}$ and $G_{p}$ denote the gate mechanisms for motion embeddings and personalization embeddings, respectively. Through gated cross-modal interaction, it dynamically modulates the fusion intensity, achieving an optimal balance among appearance preservation, textual description, and motion alignment for each entity.

\noindent \textbf{Personalization and motion injection.} 
As shown in Figure~\ref{fig:pipeline}, the motion conditions are introduced by the motion-guidance fuser, which employs an adaptive layer normalization to modulate the visual tokens. The concatenation of motion-normalized visual tokens and text tokens then undergoes 3D full attention to capture fine-grained relationships between textual descriptions and visual elements. To preserve the original foundational model knowledge, we apply additional cross-attention to facilitate interaction between the personalization embeddings and the visual tokens. In this manner, we inject the three modal conditions at different locations within the DiT blocks, achieving joint controllable video generation for multiple entities.

\noindent \textbf{Contrastive Loss.}
We also introduce a dual-branch contrastive learning framework to strengthen cross-modal alignment between entities and motion patterns. Our contrastive objective enforces three properties: (1) semantic-motion correspondence through positive entity-motion pairing, (2) inter-concept discriminability via negative entity-entity pairs, and (3) motion trajectory distinction through negative motion-motion pairs. Formally, the loss can be described as:
\begin{equation}
\mathcal{L}_{\mathrm{cont}} = 
-\sum_i \log \frac{\exp(\mathbf{p}_i^\top \mathbf{m}_i / \tau)}
{\sum_{j}\exp(\mathbf{p}_i^\top \mathbf{m}_j / \tau)} 
-\sum_j \log \frac{\exp(\mathbf{m}_j^\top \mathbf{p}_j / \tau)}
{\sum_{i}\exp(\mathbf{m}_j^\top \mathbf{p}_i / \tau)} , \tag{6}
\end{equation}
where $\tau$ denotes the temperature hyperparameter. This symmetric formulation simultaneously ensures entity-specific motion binding while expanding separation between different personalization embeddings in the joint latent space.
Finally, the specific diffusion process can be formulated as:
\begin{gather}
\mathcal{L}_{\mathrm{total}} = \mathcal{L}_{\epsilon} + \lambda \mathcal{L}_{\mathrm{cont}},  \tag{7} \\
\mathcal{L}_{\epsilon} = \mathbb{E}_{t, z_0, \epsilon} \left\| \epsilon - \epsilon_\theta(z_t, t, c, p, m) \right\|_2^2 ,\tag{8}
\end{gather}
where $\lambda$ denotes the weights of the contrastive loss. 

\begin{table*}[!t]
\centering
\label{test}
\caption{Quantitative comparison with different multi-entity video customization methods on the MSRVTT-Personalization benchmark, where bold text indicates the best result and underlined text denotes the second-best result. Tora2 demonstrates competitive customization capabilities comparable to Video Alchemist. Moreover, its customization and motion control abilities significantly surpass those of the two-stage pipeline comprising Tora with Flux.1.}
\begin{tabular}{ccccccccc}
\toprule
 &\multicolumn{4}{c}{Non-human object} & \multicolumn{4}{c}{Human}\\
\cmidrule(lr){2-5} \cmidrule(lr){6-9} 
 Method & Text-S $\uparrow$ & Vid-S $\uparrow$ &  Subj-S $\uparrow$ &  TrajError $\downarrow$ & Text-S $\uparrow$ & Vid-S $\uparrow$ &  Face-S $\uparrow$ &  TrajError $\downarrow$ \\ 
\midrule
Tora + Flux.1~\cite{zhang2024toratrajectoryorienteddiffusiontransformer,flux} & 0.254 &0.719 &0.587 &\underline{19.72} &0.265 &0.671 &0.363 &\underline{17.41}  \\ 
Video Alchemist~\cite{chen2025multisubjectopensetpersonalizationvideo} &\underline{0.268} &\textbf{0.743} &\textbf{0.626} &- & \underline{0.272} & \underline{0.694} &0.411 &-  \\ 
\textbf{Tora2~(ours)} &\textbf{0.273} &\underline{0.741} &\underline{0.615} &\textbf{17.43} &\textbf{0.274} &\textbf{0.702} &\textbf{0.419} &\textbf{13.52} \\
\bottomrule

\end{tabular}
\label{tab1}
\end{table*}

\subsection{Data collection}
Building upon Video Alchemist and Tora, we develop a two-stage data curation framework for multi-entity motion learning. The first stage rigorously filters raw videos that might negatively affect object motion training or cause ambiguity in personalization, while the second constructs precise entity-text-trajectory triplets, yielding 1.1M high-fidelity training samples through systematic annotation.

\noindent \textbf{Video Filtering.} 
Our video filtering process comprises multiple meticulous steps to ensure high-quality content suitable for personalized applications. Initially, videos exhibiting encoding errors, a resolution below 720p, or containing an excessive number of words are discarded. Subsequently, videos undergo a quality assessment based on their aesthetic score\footnote{https://github.com/christophschuhmann/improved-aesthetic-predictor} and optical flow score~\cite{DBLP:journals/pami/XuZCRYTG23}, with only those surpassing an aesthetic score of 5 and a flow score of 2 being retained. To further refine the selection, a camera motion detector\footnote{https://github.com/antiboredom/camera-motion-detector} is employed to exclude videos with significant camera movement, maintaining a zoom detection threshold between 0.4 and 0.6 and acceptable camera movement angles within $[0^{\circ}, 30^{\circ}], [150^{\circ}, 200^{\circ}], [330^{\circ}, 360^{\circ}]$.
 Finally, videos that either lack subject entity words or contain subject entity words in plural form are removed to prevent ambiguity in personalization. This comprehensive filtering protocol ensures that the remaining videos are of high quality and content clarity, providing a robust foundation for subsequent analytical processes.

\noindent \textbf{Video Annotation.} We utilize Qwen2.5-Max~\cite{yang2024qwen2} to extract nouns from the captions. Subsequently, we select three frames from the beginning, middle, and end of the video and use LISA~\cite{lai2024lisareasoningsegmentationlarge} to extract entity masks. LISA provides highly accurate segmentation results, even for similar visual appearances and textual semantics. Masks that are too large, too small, or highly fragmented are removed. We employ the center points of the three frame masks and use CoTracker3~\cite{karaev2024cotracker3simplerbetterpoint} twice to obtain more precise tracklets. 

\section{Experiments}
\subsection{Setup}
\noindent \textbf{Implementation details.} We select the open-source Tora-T2V version, based on CogVideoX-5B~\cite{yang2024cogvideox}, to initialize Tora2. In the training phase, video captions, reference conditions, and trajectory conditions—which consist of paired images, text descriptions, and trajectory points—are dropped with probabilities of 50$\%$, 33$\%$, and 33$\%$, respectively, in line with the classifier-free guidance~\cite{ho2022classifier}. The original parameters of the CogVideoX-5B model remain fixed, while the other parameters undergo fine-tuning to facilitate effective motion and appearance control. The trajectory extractor, motion-guidance fuser, Q-Former, appearance adapters, and gated self-attention layer are all subject to joint optimization. The batch size is set to 32, the learning rate to $3 \times 10^{-6}$ with AdamW~\cite{DBLP:journals/corr/KingmaB14} used as the optimizer, and the total number of training steps to 15k. Furthermore, the contrastive loss weight is set to 0.2. In the inference phase, we employ DPM~\cite{lu2022dpm} with a sampling step of 50 and a text-guidance ratio of 6.0.

\noindent \textbf{Benchmark and Metrics.} We use the MSRVTT-Personalization benchmark to evaluate our method with current approaches, which contains 2,130 video clips that are manually annotated and cover both single and multiple subjects. Additionally, for our ablation study, we manually collect and annotate 200 videos that encompass multiple concepts, from online sources. Our evaluation metrics include: 1) Text Similarity~(Text-S): Cosine similarity between CLIP~\cite{radford2021learning} text embeddings and generated frame features; 2) Video Similarity~(Vid-S): Average CLIP feature similarity between ground truth and generated videos; 3) Subject Similarity~(Subj-S): DINO-based feature alignment between reference images and generated subject segments using Grounding-DINO~\cite{liu2024grounding}; 4) Face Similarity~(Face-S): ArcFace-R100~\cite{deng2019arcface} feature consistency between reference face crops and YOLOv9-C~\cite{wang2024yolov9} detected facial regions; 5) Trajectory Error~(TrajError): Mean L1 distance between the predicted motion trajectories by CoTracker3~\cite{karaev2024cotracker3simplerbetterpoint} and ground truth. 

\subsection{Qualitative and Quantitative Analysis}
Table~\ref{tab1} shows the quantitative evaluation results. To weaken unfair comparisons caused by different fundamental models, we build another baseline that first employs the tuning-based Flux.1~\cite{flux} to generate the customized image and employs the Tora-I2V model to generate the videos.

For personalized conditions, Tora2 achieves a mean subject similarity score 1.1$\%$ lower than Video Alchemist but demonstrates a 0.8$\%$ improvement in facial attribute preservation. Despite sharing the foundational video DiT model architecture, Tora2's motion-conditioned training paradigm introduces additional complexity compared to the personalized feature learning approach of Video Alchemist. The comparable personalization performance indicates that our DPE effectively captures discriminative identity features under joint optimization with motion conditioning. Under motion-driven generation scenarios, while Video Alchemist does not provide implementation for trajectory accuracy evaluation,  this capability is outside the scope of their method's design. The Tora + Flux.1 pipeline encounters substantial degradation in subject and identity fidelity, especially in facial areas, owing to the architectural limitations of separation control. While the fine-tuned Flux.1 model successfully initializes the first frame with strong reference alignment, the subsequent frames devolve because the framework lacks continuous personalization conditioning to anchor the generative process. Notably, Tora2's joint learning strategy for fusing motion dynamics and conceptual representations consistently achieves superior generation quality metrics across both identity and motion-conditioned tasks.

\begin{figure*}
  \centering
  \includegraphics[width=0.9\textwidth]{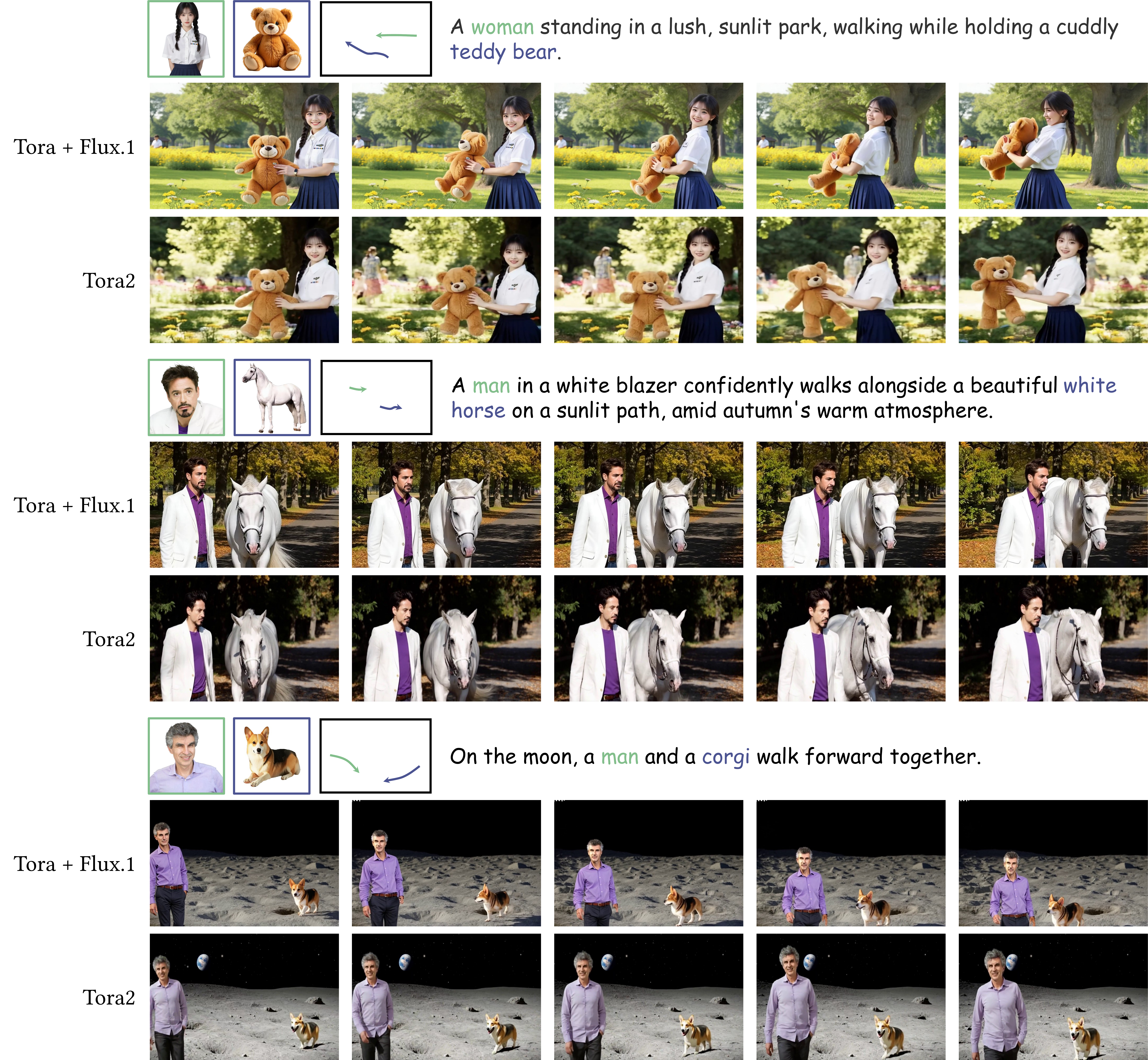}
  \caption{Qualitative comparison of appearance and motion customization for multiple entities. The depth map from the first frame of Tora2 is used as a condition for Flux.1 to ensure consistent entity positioning. The joint control of Tora2 distinctly illustrates superior performance in entity fidelity and motion smoothing.}
  \label{fig:comparison}
  \vspace{-2mm}
\end{figure*}

Figure~\ref{fig:comparison} illustrates a comparative analysis of video generations produced by contrasting approaches. Due to its closed-source status, Video Alchemist results are not included in this evaluation. Tora + Flux.1 employs a basic concatenation strategy for appearance and motion control. However, this approach results in suboptimal integration of modalities, leading to diminished entity consistency with the reference image over extended temporal sequences and excessive positional drifts in trajectory-specific regions. Such limitations compromise both spatiotemporal coherence and trajectory fidelity. In contrast, Tora2's synergistic training framework produces videos with superior photorealism, temporal smoothness, and identity preservation, achieved through effective latent space alignment between appearance and motion conditions.

\subsection{Ablation study}
For our ablation study, we utilize the annotated set of 200 videos. The Metrics for Text Similarity, Video Similarity, and Trajectory Error are averaged across both object and human entities to provide a unified illustration.

\begin{figure}[!t]
    \centering
    \includegraphics[width=0.45\textwidth]{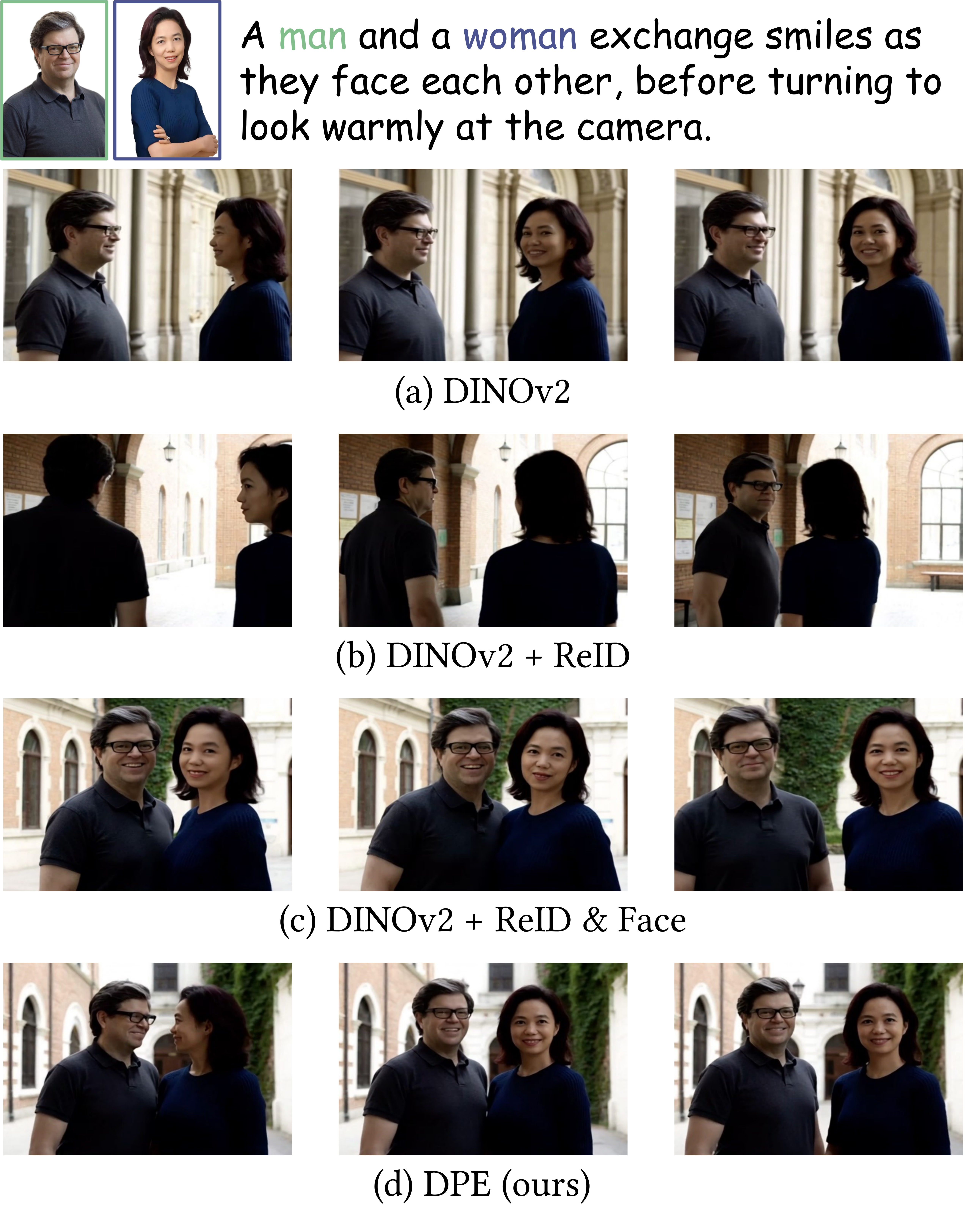}
    \caption{
        The ablation of personalization embeddings extraction. The proposed DPE adheres to the actions described in the prompts and achieves stable entity fidelity.
    }
    \label{fig:personalization}   
\end{figure}

\begin{figure}[!t]
    \centering
    \includegraphics[width=0.45\textwidth]{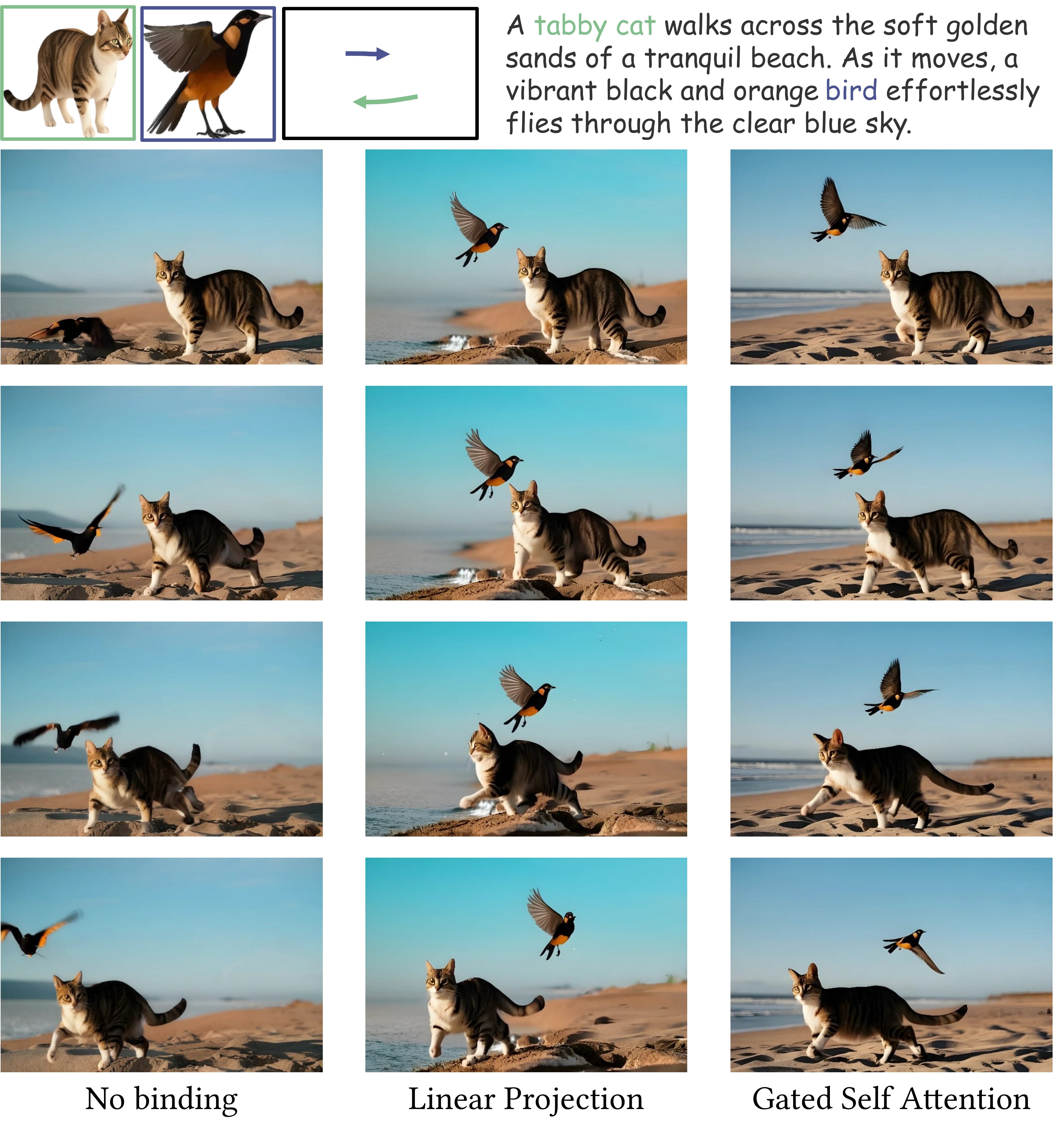}
    \caption{
        Ablation study of cross-model binding strategies. The binding mechanisms facilitate precise motion and appearance control in scenarios involving multiple entities.
    }
    \label{fig:bind}   
\end{figure}

{\fontsize {8.5pt}{8.5pt} \selectfont
\begin{table}[!t]
\centering
\setlength{\tabcolsep}{2.5pt}
\caption{Effect of different designs for personalization embedding. We exclude the motion module and the binding operator to eliminate the effects from other conditions.}
\begin{tabular}{ccccc}
\toprule
Personalization encoder            & Text-S~$\uparrow$ & Vid-S~$\uparrow$ & Subj-S~$\uparrow$ & Face-S~$\uparrow$\\ 
\midrule
DINOv2 &   0.262  &      0.717   &              0.602 & 0.389   \\
DINOv2 + ReID  &   0.253  &      0.698   &              0.599 & 0.362           \\
DINOv2 + ReID \& Face   & 0.257       &   0.725  &      0.612   &   0.404                            \\ 
\textbf{DPE~(ours)}  &   \textbf{0.266}  &      \textbf{0.733}   &       \textbf{0.621}       & \textbf{0.413}    \\
\bottomrule
\end{tabular}
\label{tab2}
\end{table}
}

\noindent \textbf{The different design for personalization embedding extraction.} 
To evaluate the effectiveness of our proposed decoupled personalization encoder, we conduct ablation experiments using different methods to extract personalization embeddings as control signals. These experiments include: (a) utilizing low-frequency global features extracted by the DINOv2 image encoder, as seen in Video Alchemist; (b) merging global features with high-frequency features from the ReID model through linear projection; (c) combining global features with decoupled high-frequency features extracted from ReID and facial recognition models via linear projection; and (d) employing our DPE, which integrates these features using a Q-Former architecture to query concatenated embeddings through learned cross-modal interactions. Figure~\ref{fig:personalization} and Table~\ref{tab2} present the qualitative and quantitative results. 

We observe that the semantic features extracted by DINOv2 can generate entities that follow text descriptions, but they struggle to transfer high-frequency details such as facial expressions. Injecting only the high-frequency discriminative features extracted by the ReID model causes significant training instability and convergence challenges, which substantially degrade visual fidelity and adherence to textual instructions when paired with the DINOv2 baseline. The decoupled strategy significantly boosts subject fidelity, making generated content visually resemble the reference entity. However, insufficient token-level interaction via MLP projection hampers the model's ability to accurately follow textual instructions, as evident in the (c) in Figure~\ref{fig:personalization} where the entities fail to look at each other as instructed.  Our proposed method achieves the best results across all metrics. This comparison confirms the necessity of combining different types of features for open-set personalization. It also demonstrates that explicit query-attend mechanisms substantially outperform naive feature concatenation in achieving better alignment within the diffusion transformer context.

{\fontsize {8pt}{8pt} \selectfont
\begin{table}[!t]\
\centering
\setlength{\tabcolsep}{2.5pt}
\caption{Quantitative comparison of binding strategy design choices. Our gated self-attention mechanism delivers superior performance.}
\begin{tabular}{cccccc}
\toprule
Binding strategy           & Text-S~$\uparrow$ & Vid-S~$\uparrow$ & Subj-S~$\uparrow$ & Face-S~$\uparrow$ & TrajError~$\downarrow$ \\ 
\midrule
No Binding &   0.258  &      0.699   &              0.589 & 0.387 & 29.95   \\
Linear Projection   &   \textbf{0.261}  &      0.708   &              0.599 & 0.391 & 20.74          \\
Gated self-attention            &   0.259  &      \textbf{0.718}   &              \textbf{0.604} & \textbf{0.395} & \textbf{17.31}              \\ 
\bottomrule
\end{tabular}
\label{tab3}
\end{table}
}

\noindent \textbf{The effect of binding different modal features.} 
We validate our entity-binding mechanism, which associates visual elements with corresponding motion trajectories and textual entities, by conducting comparative experiments across three architectural variants: (a) direct injection of personalization embeddings into the DiT blocks; (b) concatenation of personalization embeddings, motion embeddings, and textual features along the channel dimension followed by linear projection; (c) our gated self-attention mechanism that dynamically regulates feature interactions across three modalities through learnable attention gates.

Table~\ref{tab3} illustrates that the baseline architecture lacking explicit entity binding exhibits significant limitations, showing an additional 12.6-pixel offset in trajectory alignment and a 1.5$\%$ reduction in subject similarity compared to our proposed method. This performance disparity manifests visually through erroneous associations between specified motion paths and irrelevant background elements, especially when processing prompts containing multiple entities. Figure~\ref{fig:bind} provides qualitative comparison results, clearly indicating that in the absence of a binding strategy, the trajectory intended for the bird incorrectly results in moving the camera to the right, failing to control the bird as intended. In the linear projection variant, although it successfully associates trajectories, entities, and words, this relatively simple interaction does not represent the optimal choice for multi-entity association. Compared to our method, there is a decrease of 1.0$\%$ in video similarity, as evidenced by deformation observed in both the cat and bird in the latter part of the video. Our gated attention architecture emerges as the optimal solution, effectively binding the entity with the corresponding trajectory while maintaining fidelity.

\noindent \textbf{The effect of contrastive loss.} Table~\ref{tab4} quantitatively validates our contrastive learning strategy. By optimizing the relative distances between personalization embeddings and motion patterns in the latent space, it ensures precise disentanglement of multiple entities' attributes while facilitating their harmonious integration within provided motion conditions, boosting trajectory adherence accuracy by reducing the distance by approximately 3.2 pixels offset while increasing about 1.0$\%$ identity preservation fidelity.

{\fontsize {8.5pt}{8.5pt} \selectfont
\begin{table}[!t]
\centering
\setlength{\tabcolsep}{2.5pt}
\caption{Ablation study on contrastive learning. We incorporate the DPE and binding mechanism as the baseline.}
\begin{tabular}{cccccc}
\toprule
           & Text-S~$\uparrow$ & Vid-S~$\uparrow$ & Subj-S~$\uparrow$ & Face-S~$\uparrow$ & TrajError~$\downarrow$ \\ 
\midrule
baseline     &   0.259  &      0.718   &              0.604 & 0.395 & 17.31    \\
baseline + CL   &   \textbf{0.261}  &      \textbf{0.723}   &              \textbf{0.613} & \textbf{0.404} & \textbf{14.16}          \\           
\bottomrule
\end{tabular}
\label{tab4}
\end{table}
}

\noindent \textbf{The ablation of the injection order for motion and personalization embeddings}. The injection method for motion and personalization embeddings via adaptive layer normalization and cross-attention has proven to be the most effective. We adhere to these advanced designs while exploring the injection order of these features. The findings, detailed in Table~\ref{tab5}, suggest that injecting personalization embeddings early slightly reduces motion tracking fidelity. This occurs because early activation of the cross-attention mechanism disproportionately amplifies stylistic identity cues, thereby dampening the propagation of motion trajectory semantics throughout the diffusion process. As a result, we have selected the motion-text-personalization injection order as our final architecture. This arrangement ensures a nuanced signal flow: motion embeddings initially establish temporal dynamics, followed by text-based conditioning to maintain semantic integrity, and finally, personalization embeddings enhance appearance traits without sacrificing trajectory precision.

{\fontsize {8.5pt}{8.5pt} \selectfont
\begin{table}[!t]
\centering
\caption{Ablation of the injection order. The "P" denotes the appearance conditions, and "M" represents the motion conditions. }
\begin{tabular}{cccccc}
\toprule
  Order         & Text-S~$\uparrow$ & Vid-S~$\uparrow$ & Subj-S~$\uparrow$ & Face-S~$\uparrow$ & TrajError~$\downarrow$ \\ 
\midrule
P $\rightarrow$ M     &   0.253  &      0.714   &              0.611 & \textbf{0.409} & 19.23    \\
M $\rightarrow$ P   &   \textbf{0.261}  &      \textbf{0.723}   &              \textbf{0.613} & 0.404 & \textbf{14.16}          \\           
\bottomrule
\end{tabular}
\label{tab5}
\end{table}
}

\section{Conclusion}
In this paper, we introduce Tora2, a unified controllable video generation framework that effectively customizes multiple entities with motion trajectory control. Tora2 utilizes a decoupled personalization extractor to achieve open-set personalization embedding extraction by merging semantic features and high-frequency intrinsic features through a learned Q-Former framework. Additionally, it employs a novel binding strategy to associate visual entities with corresponding motion trajectories and textual words, ensuring coherent control across different entities. Extensive experiments demonstrate that Tora2 achieves competitive performance with state-of-the-art customization methods while introducing advanced motion trajectory control. These findings significantly enhance the existing capabilities of controllable video generation processes. 

\bibliographystyle{ACM-Reference-Format}
\bibliography{camera-ready}


\begin{thebibliography}{53}


\ifx \showCODEN    \undefined \def \showCODEN     #1{\unskip}     \fi
\ifx \showISBNx    \undefined \def \showISBNx     #1{\unskip}     \fi
\ifx \showISBNxiii \undefined \def \showISBNxiii  #1{\unskip}     \fi
\ifx \showISSN     \undefined \def \showISSN      #1{\unskip}     \fi
\ifx \showLCCN     \undefined \def \showLCCN      #1{\unskip}     \fi
\ifx \shownote     \undefined \def \shownote      #1{#1}          \fi
\ifx \showarticletitle \undefined \def \showarticletitle #1{#1}   \fi
\ifx \showURL      \undefined \def \showURL       {\relax}        \fi
\providecommand\bibfield[2]{#2}
\providecommand\bibinfo[2]{#2}
\providecommand\natexlab[1]{#1}
\providecommand\showeprint[2][]{arXiv:#2}

\bibitem[Bao et~al\mbox{.}(2024)]%
        {DBLP:journals/corr/abs-2405-04233}
\bibfield{author}{\bibinfo{person}{Fan Bao}, \bibinfo{person}{Chendong Xiang}, \bibinfo{person}{Gang Yue}, \bibinfo{person}{Guande He}, \bibinfo{person}{Hongzhou Zhu}, \bibinfo{person}{Kaiwen Zheng}, \bibinfo{person}{Min Zhao}, \bibinfo{person}{Shilong Liu}, \bibinfo{person}{Yaole Wang}, {and} \bibinfo{person}{Jun Zhu}.} \bibinfo{year}{2024}\natexlab{}.
\newblock \bibinfo{title}{Vidu: a Highly Consistent, Dynamic and Skilled Text-to-Video Generator with Diffusion Models}.
\newblock
\showeprint{2405.04233}


\bibitem[Blattmann et~al\mbox{.}(2023)]%
        {DBLP:journals/corr/abs-2311-15127}
\bibfield{author}{\bibinfo{person}{Andreas Blattmann}, \bibinfo{person}{Tim Dockhorn}, \bibinfo{person}{Sumith Kulal}, \bibinfo{person}{Daniel Mendelevitch}, \bibinfo{person}{Maciej Kilian}, \bibinfo{person}{Dominik Lorenz}, \bibinfo{person}{Yam Levi}, \bibinfo{person}{Zion English}, \bibinfo{person}{Vikram Voleti}, \bibinfo{person}{Adam Letts}, \bibinfo{person}{Varun Jampani}, {and} \bibinfo{person}{Robin Rombach}.} \bibinfo{year}{2023}\natexlab{}.
\newblock \bibinfo{title}{Stable Video Diffusion: Scaling Latent Video Diffusion Models to Large Datasets}.
\newblock
\showeprint{2311.15127}


\bibitem[Brooks et~al\mbox{.}(2024)]%
        {sora2024}
\bibfield{author}{\bibinfo{person}{Tim Brooks}, \bibinfo{person}{Bill Peebles}, \bibinfo{person}{Connor Holmes}, \bibinfo{person}{Will DePue}, \bibinfo{person}{Yufei Guo}, \bibinfo{person}{Li Jing}, \bibinfo{person}{David Schnurr}, \bibinfo{person}{Joe Taylor}, \bibinfo{person}{Troy Luhman}, \bibinfo{person}{Eric Luhman}, \bibinfo{person}{Clarence Ng}, \bibinfo{person}{Ricky Wang}, {and} \bibinfo{person}{Aditya Ramesh}.} \bibinfo{year}{2024}\natexlab{}.
\newblock \bibinfo{title}{Video generation models as world simulators}.
\newblock \bibinfo{howpublished}{\url{https://openai.com/research/video-generation-models-as-world-simulators}}.
\newblock


\bibitem[Chen et~al\mbox{.}(2023)]%
        {Chen2023VideoCrafter1OD}
\bibfield{author}{\bibinfo{person}{Haoxin Chen}, \bibinfo{person}{Menghan Xia}, \bibinfo{person}{Yin-Yin He}, \bibinfo{person}{Yong Zhang}, \bibinfo{person}{Xiaodong Cun}, \bibinfo{person}{Shaoshu Yang}, \bibinfo{person}{Jinbo Xing}, \bibinfo{person}{Yaofang Liu}, \bibinfo{person}{Qifeng Chen}, \bibinfo{person}{Xintao Wang}, \bibinfo{person}{Chao-Liang Weng}, {and} \bibinfo{person}{Ying Shan}.} \bibinfo{year}{2023}\natexlab{}.
\newblock \bibinfo{title}{{VideoCrafter1}: Open Diffusion Models for High-Quality Video Generation}.
\newblock
\showeprint{2310.19512}


\bibitem[Chen et~al\mbox{.}(2025)]%
        {chen2025multisubjectopensetpersonalizationvideo}
\bibfield{author}{\bibinfo{person}{Tsai-Shien Chen}, \bibinfo{person}{Aliaksandr Siarohin}, \bibinfo{person}{Willi Menapace}, \bibinfo{person}{Yuwei Fang}, \bibinfo{person}{Kwot~Sin Lee}, \bibinfo{person}{Ivan Skorokhodov}, \bibinfo{person}{Kfir Aberman}, \bibinfo{person}{Jun-Yan Zhu}, \bibinfo{person}{Ming-Hsuan Yang}, {and} \bibinfo{person}{Sergey Tulyakov}.} \bibinfo{year}{2025}\natexlab{}.
\newblock \bibinfo{title}{Multi-subject Open-set Personalization in Video Generation}.
\newblock
\showeprint[arxiv]{2501.06187}~[cs.CV]


\bibitem[Dai et~al\mbox{.}(2023)]%
        {DBLP:journals/corr/abs-2311-12886}
\bibfield{author}{\bibinfo{person}{Zuozhuo Dai}, \bibinfo{person}{Zhenghao Zhang}, \bibinfo{person}{Yao Yao}, \bibinfo{person}{Bingxue Qiu}, \bibinfo{person}{Siyu Zhu}, \bibinfo{person}{Long Qin}, {and} \bibinfo{person}{Weizhi Wang}.} \bibinfo{year}{2023}\natexlab{}.
\newblock \bibinfo{title}{Fine-Grained Open Domain Image Animation with Motion Guidance}.
\newblock
\showeprint{2311.12886}


\bibitem[Deng et~al\mbox{.}(2019)]%
        {deng2019arcface}
\bibfield{author}{\bibinfo{person}{Jiankang Deng}, \bibinfo{person}{Jia Guo}, \bibinfo{person}{Niannan Xue}, {and} \bibinfo{person}{Stefanos Zafeiriou}.} \bibinfo{year}{2019}\natexlab{}.
\newblock \showarticletitle{ArcFace: Additive Angular Margin Loss for Deep Face Recognition}. In \bibinfo{booktitle}{\emph{IEEE Conf. Comput. Vis. Pattern Recog.}} \bibinfo{publisher}{Computer Vision Foundation / {IEEE}}, \bibinfo{pages}{4690--4699}.
\newblock
\href{https://doi.org/10.1109/CVPR.2019.00482}{doi:\nolinkurl{10.1109/CVPR.2019.00482}}


\bibitem[Guo et~al\mbox{.}(2023)]%
        {guo2023animatediff}
\bibfield{author}{\bibinfo{person}{Yuwei Guo}, \bibinfo{person}{Ceyuan Yang}, \bibinfo{person}{Anyi Rao}, \bibinfo{person}{Yaohui Wang}, \bibinfo{person}{Yu Qiao}, \bibinfo{person}{Dahua Lin}, {and} \bibinfo{person}{Bo Dai}.} \bibinfo{year}{2023}\natexlab{}.
\newblock \bibinfo{title}{Animatediff: Animate your personalized text-to-image diffusion models without specific tuning}.
\newblock
\showeprint{2307.04725}


\bibitem[He et~al\mbox{.}(2024)]%
        {he2024idanimatorzeroshotidentitypreservinghuman}
\bibfield{author}{\bibinfo{person}{Xuanhua He}, \bibinfo{person}{Quande Liu}, \bibinfo{person}{Shengju Qian}, \bibinfo{person}{Xin Wang}, \bibinfo{person}{Tao Hu}, \bibinfo{person}{Ke Cao}, \bibinfo{person}{Keyu Yan}, {and} \bibinfo{person}{Jie Zhang}.} \bibinfo{year}{2024}\natexlab{}.
\newblock \bibinfo{title}{ID-Animator: Zero-Shot Identity-Preserving Human Video Generation}.
\newblock
\showeprint[arxiv]{2404.15275}~[cs.CV]


\bibitem[Ho and Salimans(2022)]%
        {ho2022classifier}
\bibfield{author}{\bibinfo{person}{Jonathan Ho} {and} \bibinfo{person}{Tim Salimans}.} \bibinfo{year}{2022}\natexlab{}.
\newblock \bibinfo{title}{Classifier-free diffusion guidance}.
\newblock
\showeprint{2207.12598}


\bibitem[Huang and Belongie(2017)]%
        {huang2017arbitrary}
\bibfield{author}{\bibinfo{person}{Xun Huang} {and} \bibinfo{person}{Serge~J. Belongie}.} \bibinfo{year}{2017}\natexlab{}.
\newblock \showarticletitle{Arbitrary Style Transfer in Real-Time with Adaptive Instance Normalization}. In \bibinfo{booktitle}{\emph{Int. Conf. Comput. Vis.}} \bibinfo{publisher}{{IEEE} Computer Society}, \bibinfo{pages}{1510--1519}.
\newblock
\href{https://doi.org/10.1109/ICCV.2017.167}{doi:\nolinkurl{10.1109/ICCV.2017.167}}


\bibitem[Huang et~al\mbox{.}(2025)]%
        {huang2025conceptmastermulticonceptvideocustomization}
\bibfield{author}{\bibinfo{person}{Yuzhou Huang}, \bibinfo{person}{Ziyang Yuan}, \bibinfo{person}{Quande Liu}, \bibinfo{person}{Qiulin Wang}, \bibinfo{person}{Xintao Wang}, \bibinfo{person}{Ruimao Zhang}, \bibinfo{person}{Pengfei Wan}, \bibinfo{person}{Di Zhang}, {and} \bibinfo{person}{Kun Gai}.} \bibinfo{year}{2025}\natexlab{}.
\newblock \bibinfo{title}{ConceptMaster: Multi-Concept Video Customization on Diffusion Transformer Models Without Test-Time Tuning}.
\newblock
\showeprint[arxiv]{2501.04698}~[cs.CV]


\bibitem[Jiang et~al\mbox{.}(2024)]%
        {jiang2023videoboothdiffusionbasedvideogeneration}
\bibfield{author}{\bibinfo{person}{Yuming Jiang}, \bibinfo{person}{Tianxing Wu}, \bibinfo{person}{Shuai Yang}, \bibinfo{person}{Chenyang Si}, \bibinfo{person}{Dahua Lin}, \bibinfo{person}{Yu Qiao}, \bibinfo{person}{Chen~Change Loy}, {and} \bibinfo{person}{Ziwei Liu}.} \bibinfo{year}{2024}\natexlab{}.
\newblock \showarticletitle{{VideoBooth}: Diffusion-based Video Generation with Image Prompts}. In \bibinfo{booktitle}{\emph{IEEE Conf. Comput. Vis. Pattern Recog.}} \bibinfo{publisher}{{IEEE}}, \bibinfo{pages}{6689--6700}.
\newblock
\href{https://doi.org/10.1109/CVPR52733.2024.00639}{doi:\nolinkurl{10.1109/CVPR52733.2024.00639}}


\bibitem[Karaev et~al\mbox{.}(2024)]%
        {karaev2024cotracker3simplerbetterpoint}
\bibfield{author}{\bibinfo{person}{Nikita Karaev}, \bibinfo{person}{Iurii Makarov}, \bibinfo{person}{Jianyuan Wang}, \bibinfo{person}{Natalia Neverova}, \bibinfo{person}{Andrea Vedaldi}, {and} \bibinfo{person}{Christian Rupprecht}.} \bibinfo{year}{2024}\natexlab{}.
\newblock \bibinfo{title}{CoTracker3: Simpler and Better Point Tracking by Pseudo-Labelling Real Videos}.
\newblock
\showeprint[arxiv]{2410.11831}~[cs.CV]


\bibitem[Kingma and Ba(2015)]%
        {DBLP:journals/corr/KingmaB14}
\bibfield{author}{\bibinfo{person}{Diederik~P. Kingma} {and} \bibinfo{person}{Jimmy Ba}.} \bibinfo{year}{2015}\natexlab{}.
\newblock \showarticletitle{Adam: {A} Method for Stochastic Optimization}. In \bibinfo{booktitle}{\emph{Int. Conf. Learn. Represent.}}, \bibfield{editor}{\bibinfo{person}{Yoshua Bengio} {and} \bibinfo{person}{Yann LeCun}} (Eds.).
\newblock


\bibitem[Kong et~al\mbox{.}(2025)]%
        {kong2025hunyuanvideosystematicframeworklarge}
\bibfield{author}{\bibinfo{person}{Weijie Kong}, \bibinfo{person}{Qi Tian}, \bibinfo{person}{Zijian Zhang}, \bibinfo{person}{Rox Min}, \bibinfo{person}{Zuozhuo Dai}, \bibinfo{person}{Jin Zhou}, \bibinfo{person}{Jiangfeng Xiong}, \bibinfo{person}{Xin Li}, \bibinfo{person}{Bo Wu}, \bibinfo{person}{Jianwei Zhang}, \bibinfo{person}{Kathrina Wu}, \bibinfo{person}{Qin Lin}, \bibinfo{person}{Junkun Yuan}, \bibinfo{person}{Yanxin Long}, \bibinfo{person}{Aladdin Wang}, \bibinfo{person}{Andong Wang}, \bibinfo{person}{Changlin Li}, \bibinfo{person}{Duojun Huang}, \bibinfo{person}{Fang Yang}, \bibinfo{person}{Hao Tan}, \bibinfo{person}{Hongmei Wang}, \bibinfo{person}{Jacob Song}, \bibinfo{person}{Jiawang Bai}, \bibinfo{person}{Jianbing Wu}, \bibinfo{person}{Jinbao Xue}, \bibinfo{person}{Joey Wang}, \bibinfo{person}{Kai Wang}, \bibinfo{person}{Mengyang Liu}, \bibinfo{person}{Pengyu Li}, \bibinfo{person}{Shuai Li}, \bibinfo{person}{Weiyan Wang}, \bibinfo{person}{Wenqing Yu}, \bibinfo{person}{Xinchi Deng},
  \bibinfo{person}{Yang Li}, \bibinfo{person}{Yi Chen}, \bibinfo{person}{Yutao Cui}, \bibinfo{person}{Yuanbo Peng}, \bibinfo{person}{Zhentao Yu}, \bibinfo{person}{Zhiyu He}, \bibinfo{person}{Zhiyong Xu}, \bibinfo{person}{Zixiang Zhou}, \bibinfo{person}{Zunnan Xu}, \bibinfo{person}{Yangyu Tao}, \bibinfo{person}{Qinglin Lu}, \bibinfo{person}{Songtao Liu}, \bibinfo{person}{Dax Zhou}, \bibinfo{person}{Hongfa Wang}, \bibinfo{person}{Yong Yang}, \bibinfo{person}{Di Wang}, \bibinfo{person}{Yuhong Liu}, \bibinfo{person}{Jie Jiang}, {and} \bibinfo{person}{Caesar Zhong}.} \bibinfo{year}{2025}\natexlab{}.
\newblock \bibinfo{title}{HunyuanVideo: A Systematic Framework For Large Video Generative Models}.
\newblock
\showeprint[arxiv]{2412.03603}~[cs.CV]


\bibitem[Labs.({[n.\,d.]})]%
        {flux}
\bibfield{author}{\bibinfo{person}{Shakker Labs.}} \bibinfo{year}{[n.\,d.]}\natexlab{}.
\newblock \bibinfo{title}{Flux.1-dev-controlnet-union-pro}.
\newblock \bibinfo{howpublished}{Accessed 2024 [Online]}.
\newblock
\urldef\tempurl%
\url{https://huggingface.co/Shakker-Labs/FLUX.1-dev-ControlNet-Union-Pro}
\showURL{%
\tempurl}


\bibitem[Lai et~al\mbox{.}(2024)]%
        {lai2024lisareasoningsegmentationlarge}
\bibfield{author}{\bibinfo{person}{Xin Lai}, \bibinfo{person}{Zhuotao Tian}, \bibinfo{person}{Yukang Chen}, \bibinfo{person}{Yanwei Li}, \bibinfo{person}{Yuhui Yuan}, \bibinfo{person}{Shu Liu}, {and} \bibinfo{person}{Jiaya Jia}.} \bibinfo{year}{2024}\natexlab{}.
\newblock \showarticletitle{{LISA:} Reasoning Segmentation via Large Language Model}. In \bibinfo{booktitle}{\emph{IEEE Conf. Comput. Vis. Pattern Recog.}} \bibinfo{publisher}{{IEEE}}, \bibinfo{pages}{9579--9589}.
\newblock
\href{https://doi.org/10.1109/CVPR52733.2024.00915}{doi:\nolinkurl{10.1109/CVPR52733.2024.00915}}


\bibitem[Li et~al\mbox{.}(2022)]%
        {li2022blipbootstrappinglanguageimagepretraining}
\bibfield{author}{\bibinfo{person}{Junnan Li}, \bibinfo{person}{Dongxu Li}, \bibinfo{person}{Caiming Xiong}, {and} \bibinfo{person}{Steven C.~H. Hoi}.} \bibinfo{year}{2022}\natexlab{}.
\newblock \showarticletitle{{BLIP:} Bootstrapping Language-Image Pre-training for Unified Vision-Language Understanding and Generation}. In \bibinfo{booktitle}{\emph{Int. Conf. Mach. Learn.}} \emph{(\bibinfo{series}{Proceedings of Machine Learning Research}, Vol.~\bibinfo{volume}{162})}, \bibfield{editor}{\bibinfo{person}{Kamalika Chaudhuri}, \bibinfo{person}{Stefanie Jegelka}, \bibinfo{person}{Le~Song}, \bibinfo{person}{Csaba Szepesv{\'{a}}ri}, \bibinfo{person}{Gang Niu}, {and} \bibinfo{person}{Sivan Sabato}} (Eds.). \bibinfo{publisher}{{PMLR}}, \bibinfo{pages}{12888--12900}.
\newblock
\urldef\tempurl%
\url{https://proceedings.mlr.press/v162/li22n.html}
\showURL{%
\tempurl}


\bibitem[Liu et~al\mbox{.}(2024)]%
        {liu2024grounding}
\bibfield{author}{\bibinfo{person}{Shilong Liu}, \bibinfo{person}{Zhaoyang Zeng}, \bibinfo{person}{Tianhe Ren}, \bibinfo{person}{Feng Li}, \bibinfo{person}{Hao Zhang}, \bibinfo{person}{Jie Yang}, \bibinfo{person}{Qing Jiang}, \bibinfo{person}{Chunyuan Li}, \bibinfo{person}{Jianwei Yang}, \bibinfo{person}{Hang Su}, \bibinfo{person}{Jun Zhu}, {and} \bibinfo{person}{Lei Zhang}.} \bibinfo{year}{2024}\natexlab{}.
\newblock \showarticletitle{Grounding {DINO:} Marrying {DINO} with Grounded Pre-training for Open-Set Object Detection}. In \bibinfo{booktitle}{\emph{Eur. Conf. Comput. Vis.}} \emph{(\bibinfo{series}{Lecture Notes in Computer Science}, Vol.~\bibinfo{volume}{15105})}, \bibfield{editor}{\bibinfo{person}{Ales Leonardis}, \bibinfo{person}{Elisa Ricci}, \bibinfo{person}{Stefan Roth}, \bibinfo{person}{Olga Russakovsky}, \bibinfo{person}{Torsten Sattler}, {and} \bibinfo{person}{G{\"{u}}l Varol}} (Eds.). \bibinfo{publisher}{Springer}, \bibinfo{pages}{38--55}.
\newblock
\href{https://doi.org/10.1007/978-3-031-72970-6\_3}{doi:\nolinkurl{10.1007/978-3-031-72970-6\_3}}


\bibitem[Liu et~al\mbox{.}(2022)]%
        {liu2022opening}
\bibfield{author}{\bibinfo{person}{Yang Liu}, \bibinfo{person}{Idil~Esen Zulfikar}, \bibinfo{person}{Jonathon Luiten}, \bibinfo{person}{Achal Dave}, \bibinfo{person}{Deva Ramanan}, \bibinfo{person}{Bastian Leibe}, \bibinfo{person}{Aljosa Osep}, {and} \bibinfo{person}{Laura Leal{-}Taix{\'{e}}}.} \bibinfo{year}{2022}\natexlab{}.
\newblock \showarticletitle{Opening up Open World Tracking}. In \bibinfo{booktitle}{\emph{IEEE Conf. Comput. Vis. Pattern Recog.}} \bibinfo{publisher}{{IEEE}}, \bibinfo{pages}{19023--19033}.
\newblock
\href{https://doi.org/10.1109/CVPR52688.2022.01846}{doi:\nolinkurl{10.1109/CVPR52688.2022.01846}}


\bibitem[Lu et~al\mbox{.}(2022)]%
        {lu2022dpm}
\bibfield{author}{\bibinfo{person}{Cheng Lu}, \bibinfo{person}{Yuhao Zhou}, \bibinfo{person}{Fan Bao}, \bibinfo{person}{Jianfei Chen}, \bibinfo{person}{Chongxuan Li}, {and} \bibinfo{person}{Jun Zhu}.} \bibinfo{year}{2022}\natexlab{}.
\newblock \showarticletitle{{DPM-Solver}: {A} Fast {ODE} Solver for Diffusion Probabilistic Model Sampling in Around 10 Steps}. In \bibinfo{booktitle}{\emph{Adv. Neural Inform. Process. Syst.}}, \bibfield{editor}{\bibinfo{person}{Sanmi Koyejo}, \bibinfo{person}{S.~Mohamed}, \bibinfo{person}{A.~Agarwal}, \bibinfo{person}{Danielle Belgrave}, \bibinfo{person}{K.~Cho}, {and} \bibinfo{person}{A.~Oh}} (Eds.), Vol.~\bibinfo{volume}{35}. \bibinfo{pages}{5775--5787}.
\newblock
\urldef\tempurl%
\url{http://papers.nips.cc/paper\_files/paper/2022/hash/260a14acce2a89dad36adc8eefe7c59e-Abstract-Conference.html}
\showURL{%
\tempurl}


\bibitem[Ma et~al\mbox{.}(2024)]%
        {DBLP:journals/corr/abs-2401-00896}
\bibfield{author}{\bibinfo{person}{Wan{-}Duo~Kurt Ma}, \bibinfo{person}{John~P. Lewis}, {and} \bibinfo{person}{W.~Bastiaan Kleijn}.} \bibinfo{year}{2024}\natexlab{}.
\newblock \showarticletitle{TrailBlazer: Trajectory Control for Diffusion-Based Video Generation}. In \bibinfo{booktitle}{\emph{{SIGGRAPH} Asia}}, \bibfield{editor}{\bibinfo{person}{Takeo Igarashi}, \bibinfo{person}{Ariel Shamir}, {and} \bibinfo{person}{Hao~(Richard) Zhang}} (Eds.). \bibinfo{publisher}{{ACM}}, \bibinfo{pages}{97:1--97:11}.
\newblock
\href{https://doi.org/10.1145/3680528.3687652}{doi:\nolinkurl{10.1145/3680528.3687652}}


\bibitem[Mallya et~al\mbox{.}(2022)]%
        {mallya2022implicit}
\bibfield{author}{\bibinfo{person}{Arun Mallya}, \bibinfo{person}{Ting{-}Chun Wang}, {and} \bibinfo{person}{Ming{-}Yu Liu}.} \bibinfo{year}{2022}\natexlab{}.
\newblock \showarticletitle{Implicit Warping for Animation with Image Sets}. In \bibinfo{booktitle}{\emph{Adv. Neural Inform. Process. Syst.}}, \bibfield{editor}{\bibinfo{person}{Sanmi Koyejo}, \bibinfo{person}{S.~Mohamed}, \bibinfo{person}{A.~Agarwal}, \bibinfo{person}{Danielle Belgrave}, \bibinfo{person}{K.~Cho}, {and} \bibinfo{person}{A.~Oh}} (Eds.). \bibinfo{pages}{22438--22450}.
\newblock
\urldef\tempurl%
\url{http://papers.nips.cc/paper\_files/paper/2022/hash/8cb31912235561112339f04903657f72-Abstract-Conference.html}
\showURL{%
\tempurl}


\bibitem[Oquab et~al\mbox{.}(2024)]%
        {oquab2024dinov2learningrobustvisual}
\bibfield{author}{\bibinfo{person}{Maxime Oquab}, \bibinfo{person}{Timoth{\'{e}}e Darcet}, \bibinfo{person}{Th{\'{e}}o Moutakanni}, \bibinfo{person}{Huy~V. Vo}, \bibinfo{person}{Marc Szafraniec}, \bibinfo{person}{Vasil Khalidov}, \bibinfo{person}{Pierre Fernandez}, \bibinfo{person}{Daniel Haziza}, \bibinfo{person}{Francisco Massa}, \bibinfo{person}{Alaaeldin El{-}Nouby}, \bibinfo{person}{Mido Assran}, \bibinfo{person}{Nicolas Ballas}, \bibinfo{person}{Wojciech Galuba}, \bibinfo{person}{Russell Howes}, \bibinfo{person}{Po{-}Yao Huang}, \bibinfo{person}{Shang{-}Wen Li}, \bibinfo{person}{Ishan Misra}, \bibinfo{person}{Michael Rabbat}, \bibinfo{person}{Vasu Sharma}, \bibinfo{person}{Gabriel Synnaeve}, \bibinfo{person}{Hu Xu}, \bibinfo{person}{Herv{\'{e}} J{\'{e}}gou}, \bibinfo{person}{Julien Mairal}, \bibinfo{person}{Patrick Labatut}, \bibinfo{person}{Armand Joulin}, {and} \bibinfo{person}{Piotr Bojanowski}.} \bibinfo{year}{2024}\natexlab{}.
\newblock \showarticletitle{DINOv2: Learning Robust Visual Features without Supervision}.
\newblock \bibinfo{journal}{\emph{Trans. Mach. Learn. Res.}}  \bibinfo{volume}{2024} (\bibinfo{year}{2024}).
\newblock
\urldef\tempurl%
\url{https://openreview.net/forum?id=a68SUt6zFt}
\showURL{%
\tempurl}


\bibitem[Peebles and Xie(2023)]%
        {peebles2023scalable}
\bibfield{author}{\bibinfo{person}{William Peebles} {and} \bibinfo{person}{Saining Xie}.} \bibinfo{year}{2023}\natexlab{}.
\newblock \showarticletitle{Scalable Diffusion Models with Transformers}. In \bibinfo{booktitle}{\emph{Int. Conf. Comput. Vis.}} \bibinfo{publisher}{{IEEE}}, \bibinfo{pages}{4172--4182}.
\newblock
\href{https://doi.org/10.1109/ICCV51070.2023.00387}{doi:\nolinkurl{10.1109/ICCV51070.2023.00387}}


\bibitem[Polyak et~al\mbox{.}(2024)]%
        {polyak2024movie}
\bibfield{author}{\bibinfo{person}{Adam Polyak}, \bibinfo{person}{Amit Zohar}, \bibinfo{person}{Andrew Brown}, \bibinfo{person}{Andros Tjandra}, \bibinfo{person}{Animesh Sinha}, \bibinfo{person}{Ann Lee}, \bibinfo{person}{Apoorv Vyas}, \bibinfo{person}{Bowen Shi}, \bibinfo{person}{Chih{-}Yao Ma}, \bibinfo{person}{Ching{-}Yao Chuang}, \bibinfo{person}{David Yan}, \bibinfo{person}{Dhruv Choudhary}, \bibinfo{person}{Dingkang Wang}, \bibinfo{person}{Geet Sethi}, \bibinfo{person}{Guan Pang}, \bibinfo{person}{Haoyu Ma}, \bibinfo{person}{Ishan Misra}, \bibinfo{person}{Ji Hou}, \bibinfo{person}{Jialiang Wang}, \bibinfo{person}{Kiran Jagadeesh}, \bibinfo{person}{Kunpeng Li}, \bibinfo{person}{Luxin Zhang}, \bibinfo{person}{Mannat Singh}, \bibinfo{person}{Mary Williamson}, \bibinfo{person}{Matt Le}, \bibinfo{person}{Matthew Yu}, \bibinfo{person}{Mitesh~Kumar Singh}, \bibinfo{person}{Peizhao Zhang}, \bibinfo{person}{Peter Vajda}, \bibinfo{person}{Quentin Duval}, \bibinfo{person}{Rohit Girdhar}, \bibinfo{person}{Roshan
  Sumbaly}, \bibinfo{person}{Sai~Saketh Rambhatla}, \bibinfo{person}{Sam~S. Tsai}, \bibinfo{person}{Samaneh Azadi}, \bibinfo{person}{Samyak Datta}, \bibinfo{person}{Sanyuan Chen}, \bibinfo{person}{Sean Bell}, \bibinfo{person}{Sharadh Ramaswamy}, \bibinfo{person}{Shelly Sheynin}, \bibinfo{person}{Siddharth Bhattacharya}, \bibinfo{person}{Simran Motwani}, \bibinfo{person}{Tao Xu}, \bibinfo{person}{Tianhe Li}, \bibinfo{person}{Tingbo Hou}, \bibinfo{person}{Wei{-}Ning Hsu}, \bibinfo{person}{Xi Yin}, \bibinfo{person}{Xiaoliang Dai}, \bibinfo{person}{Yaniv Taigman}, \bibinfo{person}{Yaqiao Luo}, \bibinfo{person}{Yen{-}Cheng Liu}, \bibinfo{person}{Yi{-}Chiao Wu}, \bibinfo{person}{Yue Zhao}, \bibinfo{person}{Yuval Kirstain}, \bibinfo{person}{Zecheng He}, \bibinfo{person}{Zijian He}, \bibinfo{person}{Albert Pumarola}, \bibinfo{person}{Ali~K. Thabet}, \bibinfo{person}{Artsiom Sanakoyeu}, \bibinfo{person}{Arun Mallya}, \bibinfo{person}{Baishan Guo}, \bibinfo{person}{Boris Araya}, \bibinfo{person}{Breena Kerr},
  \bibinfo{person}{Carleigh Wood}, \bibinfo{person}{Ce Liu}, \bibinfo{person}{Cen Peng}, \bibinfo{person}{Dmitry Vengertsev}, \bibinfo{person}{Edgar Sch{\"{o}}nfeld}, \bibinfo{person}{Elliot Blanchard}, \bibinfo{person}{Felix Juefei{-}Xu}, \bibinfo{person}{Fraylie Nord}, \bibinfo{person}{Jeff Liang}, \bibinfo{person}{John Hoffman}, \bibinfo{person}{Jonas Kohler}, \bibinfo{person}{Kaolin Fire}, \bibinfo{person}{Karthik Sivakumar}, \bibinfo{person}{Lawrence Chen}, \bibinfo{person}{Licheng Yu}, \bibinfo{person}{Luya Gao}, \bibinfo{person}{Markos Georgopoulos}, \bibinfo{person}{Rashel Moritz}, \bibinfo{person}{Sara~K. Sampson}, \bibinfo{person}{Shikai Li}, \bibinfo{person}{Simone Parmeggiani}, \bibinfo{person}{Steve Fine}, \bibinfo{person}{Tara Fowler}, \bibinfo{person}{Vladan Petrovic}, {and} \bibinfo{person}{Yuming Du}.} \bibinfo{year}{2024}\natexlab{}.
\newblock \bibinfo{title}{{Movie Gen}: {A} Cast of Media Foundation Models}.
\newblock
\showeprint[arXiv]{2410.13720}


\bibitem[Radford et~al\mbox{.}(2021)]%
        {radford2021learning}
\bibfield{author}{\bibinfo{person}{Alec Radford}, \bibinfo{person}{Jong~Wook Kim}, \bibinfo{person}{Chris Hallacy}, \bibinfo{person}{Aditya Ramesh}, \bibinfo{person}{Gabriel Goh}, \bibinfo{person}{Sandhini Agarwal}, \bibinfo{person}{Girish Sastry}, \bibinfo{person}{Amanda Askell}, \bibinfo{person}{Pamela Mishkin}, \bibinfo{person}{Jack Clark}, \bibinfo{person}{Gretchen Krueger}, {and} \bibinfo{person}{Ilya Sutskever}.} \bibinfo{year}{2021}\natexlab{}.
\newblock \showarticletitle{Learning Transferable Visual Models From Natural Language Supervision}. In \bibinfo{booktitle}{\emph{Int. Conf. Mach. Learn.}} \emph{(\bibinfo{series}{Proceedings of Machine Learning Research}, Vol.~\bibinfo{volume}{139})}, \bibfield{editor}{\bibinfo{person}{Marina Meila} {and} \bibinfo{person}{Tong Zhang}} (Eds.). \bibinfo{publisher}{{PMLR}}, \bibinfo{pages}{8748--8763}.
\newblock
\urldef\tempurl%
\url{http://proceedings.mlr.press/v139/radford21a.html}
\showURL{%
\tempurl}


\bibitem[Ren et~al\mbox{.}(2024)]%
        {ren2024groundedsamassemblingopenworld}
\bibfield{author}{\bibinfo{person}{Tianhe Ren}, \bibinfo{person}{Shilong Liu}, \bibinfo{person}{Ailing Zeng}, \bibinfo{person}{Jing Lin}, \bibinfo{person}{Kunchang Li}, \bibinfo{person}{He Cao}, \bibinfo{person}{Jiayu Chen}, \bibinfo{person}{Xinyu Huang}, \bibinfo{person}{Yukang Chen}, \bibinfo{person}{Feng Yan}, \bibinfo{person}{Zhaoyang Zeng}, \bibinfo{person}{Hao Zhang}, \bibinfo{person}{Feng Li}, \bibinfo{person}{Jie Yang}, \bibinfo{person}{Hongyang Li}, \bibinfo{person}{Qing Jiang}, {and} \bibinfo{person}{Lei Zhang}.} \bibinfo{year}{2024}\natexlab{}.
\newblock \bibinfo{title}{Grounded SAM: Assembling Open-World Models for Diverse Visual Tasks}.
\newblock
\showeprint[arxiv]{2401.14159}~[cs.CV]


\bibitem[Ronneberger et~al\mbox{.}(2015)]%
        {ronneberger2015u}
\bibfield{author}{\bibinfo{person}{Olaf Ronneberger}, \bibinfo{person}{Philipp Fischer}, {and} \bibinfo{person}{Thomas Brox}.} \bibinfo{year}{2015}\natexlab{}.
\newblock \showarticletitle{{U-Net}: Convolutional Networks for Biomedical Image Segmentation}. In \bibinfo{booktitle}{\emph{Medical Image Computing and Computer-Assisted Intervention}} \emph{(\bibinfo{series}{Lecture Notes in Computer Science}, Vol.~\bibinfo{volume}{9351})}, \bibfield{editor}{\bibinfo{person}{Nassir Navab}, \bibinfo{person}{Joachim Hornegger}, \bibinfo{person}{William M.~Wells III}, {and} \bibinfo{person}{Alejandro~F. Frangi}} (Eds.). \bibinfo{publisher}{Springer}, \bibinfo{pages}{234--241}.
\newblock
\href{https://doi.org/10.1007/978-3-319-24574-4\_28}{doi:\nolinkurl{10.1007/978-3-319-24574-4\_28}}


\bibitem[Ruiz et~al\mbox{.}(2023)]%
        {ruiz2023dreamboothfinetuningtexttoimage}
\bibfield{author}{\bibinfo{person}{Nataniel Ruiz}, \bibinfo{person}{Yuanzhen Li}, \bibinfo{person}{Varun Jampani}, \bibinfo{person}{Yael Pritch}, \bibinfo{person}{Michael Rubinstein}, {and} \bibinfo{person}{Kfir Aberman}.} \bibinfo{year}{2023}\natexlab{}.
\newblock \showarticletitle{DreamBooth: Fine Tuning Text-to-Image Diffusion Models for Subject-Driven Generation}. In \bibinfo{booktitle}{\emph{IEEE Conf. Comput. Vis. Pattern Recog.}} \bibinfo{publisher}{{IEEE}}, \bibinfo{pages}{22500--22510}.
\newblock
\href{https://doi.org/10.1109/CVPR52729.2023.02155}{doi:\nolinkurl{10.1109/CVPR52729.2023.02155}}


\bibitem[Vaswani et~al\mbox{.}(2017)]%
        {DBLP:conf/nips/VaswaniSPUJGKP17}
\bibfield{author}{\bibinfo{person}{Ashish Vaswani}, \bibinfo{person}{Noam Shazeer}, \bibinfo{person}{Niki Parmar}, \bibinfo{person}{Jakob Uszkoreit}, \bibinfo{person}{Llion Jones}, \bibinfo{person}{Aidan~N. Gomez}, \bibinfo{person}{Lukasz Kaiser}, {and} \bibinfo{person}{Illia Polosukhin}.} \bibinfo{year}{2017}\natexlab{}.
\newblock \showarticletitle{Attention is All you Need}. In \bibinfo{booktitle}{\emph{Adv. Neural Inform. Process. Syst.}}, \bibfield{editor}{\bibinfo{person}{Isabelle Guyon}, \bibinfo{person}{Ulrike von Luxburg}, \bibinfo{person}{Samy Bengio}, \bibinfo{person}{Hanna~M. Wallach}, \bibinfo{person}{Rob Fergus}, \bibinfo{person}{S.~V.~N. Vishwanathan}, {and} \bibinfo{person}{Roman Garnett}} (Eds.). \bibinfo{pages}{5998--6008}.
\newblock
\urldef\tempurl%
\url{https://proceedings.neurips.cc/paper/2017/hash/3f5ee243547dee91fbd053c1c4a845aa-Abstract.html}
\showURL{%
\tempurl}


\bibitem[Wang et~al\mbox{.}(2024b)]%
        {wang2024yolov9}
\bibfield{author}{\bibinfo{person}{Chien{-}Yao Wang}, \bibinfo{person}{I{-}Hau Yeh}, {and} \bibinfo{person}{Hong{-}Yuan~Mark Liao}.} \bibinfo{year}{2024}\natexlab{b}.
\newblock \showarticletitle{YOLOv9: Learning What You Want to Learn Using Programmable Gradient Information}. In \bibinfo{booktitle}{\emph{Eur. Conf. Comput. Vis.}} \emph{(\bibinfo{series}{Lecture Notes in Computer Science}, Vol.~\bibinfo{volume}{15089})}, \bibfield{editor}{\bibinfo{person}{Ales Leonardis}, \bibinfo{person}{Elisa Ricci}, \bibinfo{person}{Stefan Roth}, \bibinfo{person}{Olga Russakovsky}, \bibinfo{person}{Torsten Sattler}, {and} \bibinfo{person}{G{\"{u}}l Varol}} (Eds.). \bibinfo{publisher}{Springer}, \bibinfo{pages}{1--21}.
\newblock
\href{https://doi.org/10.1007/978-3-031-72751-1\_1}{doi:\nolinkurl{10.1007/978-3-031-72751-1\_1}}


\bibitem[Wang et~al\mbox{.}(2024a)]%
        {wang2024levitor3dtrajectoryoriented}
\bibfield{author}{\bibinfo{person}{Hanlin Wang}, \bibinfo{person}{Hao Ouyang}, \bibinfo{person}{Qiuyu Wang}, \bibinfo{person}{Wen Wang}, \bibinfo{person}{Ka~Leong Cheng}, \bibinfo{person}{Qifeng Chen}, \bibinfo{person}{Yujun Shen}, {and} \bibinfo{person}{Limin Wang}.} \bibinfo{year}{2024}\natexlab{a}.
\newblock \bibinfo{title}{LeviTor: 3D Trajectory Oriented Image-to-Video Synthesis}.
\newblock
\showeprint[arxiv]{2412.15214}~[cs.CV]


\bibitem[Wang et~al\mbox{.}(2023a)]%
        {wang2023modelscope}
\bibfield{author}{\bibinfo{person}{Jiuniu Wang}, \bibinfo{person}{Hangjie Yuan}, \bibinfo{person}{Dayou Chen}, \bibinfo{person}{Yingya Zhang}, \bibinfo{person}{Xiang Wang}, {and} \bibinfo{person}{Shiwei Zhang}.} \bibinfo{year}{2023}\natexlab{a}.
\newblock \bibinfo{title}{Modelscope text-to-video technical report}.
\newblock
\showeprint{2308.06571}


\bibitem[Wang et~al\mbox{.}(2023c)]%
        {wang2023videocomposer}
\bibfield{author}{\bibinfo{person}{Xiang Wang}, \bibinfo{person}{Hangjie Yuan}, \bibinfo{person}{Shiwei Zhang}, \bibinfo{person}{Dayou Chen}, \bibinfo{person}{Jiuniu Wang}, \bibinfo{person}{Yingya Zhang}, \bibinfo{person}{Yujun Shen}, \bibinfo{person}{Deli Zhao}, {and} \bibinfo{person}{Jingren Zhou}.} \bibinfo{year}{2023}\natexlab{c}.
\newblock \showarticletitle{{VideoComposer}: Compositional Video Synthesis with Motion Controllability}. In \bibinfo{booktitle}{\emph{Adv. Neural Inform. Process. Syst.}}, \bibfield{editor}{\bibinfo{person}{Alice Oh}, \bibinfo{person}{Tristan Naumann}, \bibinfo{person}{Amir Globerson}, \bibinfo{person}{Kate Saenko}, \bibinfo{person}{Moritz Hardt}, {and} \bibinfo{person}{Sergey Levine}} (Eds.), Vol.~\bibinfo{volume}{36}. \bibinfo{pages}{7594--7611}.
\newblock
\urldef\tempurl%
\url{http://papers.nips.cc/paper\_files/paper/2023/hash/180f6184a3458fa19c28c5483bc61877-Abstract-Conference.html}
\showURL{%
\tempurl}


\bibitem[Wang et~al\mbox{.}(2024c)]%
        {wang2024tf}
\bibfield{author}{\bibinfo{person}{Xiang Wang}, \bibinfo{person}{Shiwei Zhang}, \bibinfo{person}{Hangjie Yuan}, \bibinfo{person}{Zhiwu Qing}, \bibinfo{person}{Biao Gong}, \bibinfo{person}{Yingya Zhang}, \bibinfo{person}{Yujun Shen}, \bibinfo{person}{Changxin Gao}, {and} \bibinfo{person}{Nong Sang}.} \bibinfo{year}{2024}\natexlab{c}.
\newblock \showarticletitle{A Recipe for Scaling up Text-to-Video Generation with Text-free Videos}. In \bibinfo{booktitle}{\emph{IEEE Conf. Comput. Vis. Pattern Recog.}} \bibinfo{publisher}{{IEEE}}, \bibinfo{pages}{6572--6582}.
\newblock
\href{https://doi.org/10.1109/CVPR52733.2024.00628}{doi:\nolinkurl{10.1109/CVPR52733.2024.00628}}


\bibitem[Wang et~al\mbox{.}(2023b)]%
        {wang2023motionctrl}
\bibfield{author}{\bibinfo{person}{Zhouxia Wang}, \bibinfo{person}{Ziyang Yuan}, \bibinfo{person}{Xintao Wang}, \bibinfo{person}{Tianshui Chen}, \bibinfo{person}{Menghan Xia}, \bibinfo{person}{Ping Luo}, {and} \bibinfo{person}{Yin Shan}.} \bibinfo{year}{2023}\natexlab{b}.
\newblock \bibinfo{title}{MotionCtrl: A Unified and Flexible Motion Controller for Video Generation}.
\newblock
\showeprint{2312.03641}


\bibitem[Wei et~al\mbox{.}(2024a)]%
        {wei2023dreamvideocomposingdreamvideos}
\bibfield{author}{\bibinfo{person}{Yujie Wei}, \bibinfo{person}{Shiwei Zhang}, \bibinfo{person}{Zhiwu Qing}, \bibinfo{person}{Hangjie Yuan}, \bibinfo{person}{Zhiheng Liu}, \bibinfo{person}{Yu Liu}, \bibinfo{person}{Yingya Zhang}, \bibinfo{person}{Jingren Zhou}, {and} \bibinfo{person}{Hongming Shan}.} \bibinfo{year}{2024}\natexlab{a}.
\newblock \showarticletitle{{Dream Video}: Composing Your Dream Videos with Customized Subject and Motion}. In \bibinfo{booktitle}{\emph{IEEE Conf. Comput. Vis. Pattern Recog.}} \bibinfo{publisher}{{IEEE}}, \bibinfo{pages}{6537--6549}.
\newblock
\href{https://doi.org/10.1109/CVPR52733.2024.00625}{doi:\nolinkurl{10.1109/CVPR52733.2024.00625}}


\bibitem[Wei et~al\mbox{.}(2024b)]%
        {wei2024dreamvideo2zeroshotsubjectdrivenvideo}
\bibfield{author}{\bibinfo{person}{Yujie Wei}, \bibinfo{person}{Shiwei Zhang}, \bibinfo{person}{Hangjie Yuan}, \bibinfo{person}{Xiang Wang}, \bibinfo{person}{Haonan Qiu}, \bibinfo{person}{Rui Zhao}, \bibinfo{person}{Yutong Feng}, \bibinfo{person}{Feng Liu}, \bibinfo{person}{Zhizhong Huang}, \bibinfo{person}{Jiaxin Ye}, \bibinfo{person}{Yingya Zhang}, {and} \bibinfo{person}{Hongming Shan}.} \bibinfo{year}{2024}\natexlab{b}.
\newblock \bibinfo{title}{DreamVideo-2: Zero-Shot Subject-Driven Video Customization with Precise Motion Control}.
\newblock
\showeprint[arxiv]{2410.13830}~[cs.CV]


\bibitem[Wu et~al\mbox{.}(2024b)]%
        {wu2024motionboothmotionawarecustomizedtexttovideo}
\bibfield{author}{\bibinfo{person}{Jianzong Wu}, \bibinfo{person}{Xiangtai Li}, \bibinfo{person}{Yanhong Zeng}, \bibinfo{person}{Jiangning Zhang}, \bibinfo{person}{Qianyu Zhou}, \bibinfo{person}{Yining Li}, \bibinfo{person}{Yunhai Tong}, {and} \bibinfo{person}{Kai Chen}.} \bibinfo{year}{2024}\natexlab{b}.
\newblock \bibinfo{title}{MotionBooth: Motion-Aware Customized Text-to-Video Generation}.
\newblock
\showeprint[arxiv]{2406.17758}~[cs.CV]


\bibitem[Wu et~al\mbox{.}(2024a)]%
        {wu2024draganythingmotioncontrolusing}
\bibfield{author}{\bibinfo{person}{Weijia Wu}, \bibinfo{person}{Zhuang Li}, \bibinfo{person}{Yuchao Gu}, \bibinfo{person}{Rui Zhao}, \bibinfo{person}{Yefei He}, \bibinfo{person}{David~Junhao Zhang}, \bibinfo{person}{Mike~Zheng Shou}, \bibinfo{person}{Yan Li}, \bibinfo{person}{Tingting Gao}, {and} \bibinfo{person}{Di Zhang}.} \bibinfo{year}{2024}\natexlab{a}.
\newblock \showarticletitle{{DragAnything}: Motion Control for Anything Using Entity Representation}. In \bibinfo{booktitle}{\emph{Eur. Conf. Comput. Vis.}} \emph{(\bibinfo{series}{Lecture Notes in Computer Science}, Vol.~\bibinfo{volume}{15080})}, \bibfield{editor}{\bibinfo{person}{Ales Leonardis}, \bibinfo{person}{Elisa Ricci}, \bibinfo{person}{Stefan Roth}, \bibinfo{person}{Olga Russakovsky}, \bibinfo{person}{Torsten Sattler}, {and} \bibinfo{person}{G{\"{u}}l Varol}} (Eds.). \bibinfo{publisher}{Springer}, \bibinfo{pages}{331--348}.
\newblock
\href{https://doi.org/10.1007/978-3-031-72670-5\_19}{doi:\nolinkurl{10.1007/978-3-031-72670-5\_19}}


\bibitem[Xu et~al\mbox{.}(2023)]%
        {DBLP:journals/pami/XuZCRYTG23}
\bibfield{author}{\bibinfo{person}{Haofei Xu}, \bibinfo{person}{Jing Zhang}, \bibinfo{person}{Jianfei Cai}, \bibinfo{person}{Hamid Rezatofighi}, \bibinfo{person}{Fisher Yu}, \bibinfo{person}{Dacheng Tao}, {and} \bibinfo{person}{Andreas Geiger}.} \bibinfo{year}{2023}\natexlab{}.
\newblock \showarticletitle{Unifying Flow, Stereo and Depth Estimation}.
\newblock \bibinfo{journal}{\emph{IEEE Trans. Pattern Anal. Mach. Intell.}} \bibinfo{volume}{45}, \bibinfo{number}{11} (\bibinfo{year}{2023}), \bibinfo{pages}{13941--13958}.
\newblock
\href{https://doi.org/10.1109/TPAMI.2023.3298645}{doi:\nolinkurl{10.1109/TPAMI.2023.3298645}}


\bibitem[Yang et~al\mbox{.}(2024b)]%
        {yang2024qwen2}
\bibfield{author}{\bibinfo{person}{An Yang}, \bibinfo{person}{Baosong Yang}, \bibinfo{person}{Beichen Zhang}, \bibinfo{person}{Binyuan Hui}, \bibinfo{person}{Bo Zheng}, \bibinfo{person}{Bowen Yu}, \bibinfo{person}{Chengyuan Li}, \bibinfo{person}{Dayiheng Liu}, \bibinfo{person}{Fei Huang}, \bibinfo{person}{Haoran Wei}, {et~al\mbox{.}}} \bibinfo{year}{2024}\natexlab{b}.
\newblock \bibinfo{title}{Qwen2.5 technical report}.
\newblock
\showeprint{2412.15115}


\bibitem[Yang et~al\mbox{.}(2024a)]%
        {yang2024cogvideox}
\bibfield{author}{\bibinfo{person}{Zhuoyi Yang}, \bibinfo{person}{Jiayan Teng}, \bibinfo{person}{Wendi Zheng}, \bibinfo{person}{Ming Ding}, \bibinfo{person}{Shiyu Huang}, \bibinfo{person}{Jiazheng Xu}, \bibinfo{person}{Yuanming Yang}, \bibinfo{person}{Wenyi Hong}, \bibinfo{person}{Xiaohan Zhang}, \bibinfo{person}{Guanyu Feng}, {et~al\mbox{.}}} \bibinfo{year}{2024}\natexlab{a}.
\newblock \bibinfo{title}{{CogVideoX}: Text-to-Video Diffusion Models with An Expert Transformer}.
\newblock
\showeprint{2408.06072}


\bibitem[Ye et~al\mbox{.}(2023)]%
        {ye2023ipadaptertextcompatibleimage}
\bibfield{author}{\bibinfo{person}{Hu Ye}, \bibinfo{person}{Jun Zhang}, \bibinfo{person}{Sibo Liu}, \bibinfo{person}{Xiao Han}, {and} \bibinfo{person}{Wei Yang}.} \bibinfo{year}{2023}\natexlab{}.
\newblock \bibinfo{title}{IP-Adapter: Text Compatible Image Prompt Adapter for Text-to-Image Diffusion Models}.
\newblock
\showeprint[arxiv]{2308.06721}~[cs.CV]


\bibitem[Yin et~al\mbox{.}(2023)]%
        {yin2023dragnuwa}
\bibfield{author}{\bibinfo{person}{Shengming Yin}, \bibinfo{person}{Chenfei Wu}, \bibinfo{person}{Jian Liang}, \bibinfo{person}{Jie Shi}, \bibinfo{person}{Houqiang Li}, \bibinfo{person}{Gong Ming}, {and} \bibinfo{person}{Nan Duan}.} \bibinfo{year}{2023}\natexlab{}.
\newblock \bibinfo{title}{{DragNUWA}: Fine-grained control in video generation by integrating text, image, and trajectory}.
\newblock
\showeprint{2308.08089}


\bibitem[Yu et~al\mbox{.}(2023)]%
        {yu2023magvit}
\bibfield{author}{\bibinfo{person}{Lijun Yu}, \bibinfo{person}{Yong Cheng}, \bibinfo{person}{Kihyuk Sohn}, \bibinfo{person}{Jos{\'{e}} Lezama}, \bibinfo{person}{Han Zhang}, \bibinfo{person}{Huiwen Chang}, \bibinfo{person}{Alexander~G. Hauptmann}, \bibinfo{person}{Ming{-}Hsuan Yang}, \bibinfo{person}{Yuan Hao}, \bibinfo{person}{Irfan Essa}, {and} \bibinfo{person}{Lu Jiang}.} \bibinfo{year}{2023}\natexlab{}.
\newblock \showarticletitle{{MAGVIT:} Masked Generative Video Transformer}. In \bibinfo{booktitle}{\emph{IEEE Conf. Comput. Vis. Pattern Recog.}} \bibinfo{publisher}{{IEEE}}, \bibinfo{pages}{10459--10469}.
\newblock
\href{https://doi.org/10.1109/CVPR52729.2023.01008}{doi:\nolinkurl{10.1109/CVPR52729.2023.01008}}


\bibitem[Yuan et~al\mbox{.}(2024)]%
        {yuan2024identitypreservingtexttovideogenerationfrequency}
\bibfield{author}{\bibinfo{person}{Shenghai Yuan}, \bibinfo{person}{Jinfa Huang}, \bibinfo{person}{Xianyi He}, \bibinfo{person}{Yunyuan Ge}, \bibinfo{person}{Yujun Shi}, \bibinfo{person}{Liuhan Chen}, \bibinfo{person}{Jiebo Luo}, {and} \bibinfo{person}{Li Yuan}.} \bibinfo{year}{2024}\natexlab{}.
\newblock \bibinfo{title}{Identity-Preserving Text-to-Video Generation by Frequency Decomposition}.
\newblock
\showeprint[arxiv]{2411.17440}~[cs.CV]


\bibitem[Zhang et~al\mbox{.}(2023)]%
        {DBLP:journals/corr/abs-2311-04145}
\bibfield{author}{\bibinfo{person}{Shiwei Zhang}, \bibinfo{person}{Jiayu Wang}, \bibinfo{person}{Yingya Zhang}, \bibinfo{person}{Kang Zhao}, \bibinfo{person}{Hangjie Yuan}, \bibinfo{person}{Zhiwu Qin}, \bibinfo{person}{Xiang Wang}, \bibinfo{person}{Deli Zhao}, {and} \bibinfo{person}{Jingren Zhou}.} \bibinfo{year}{2023}\natexlab{}.
\newblock \bibinfo{title}{{I2VGen-XL}: High-Quality Image-to-Video Synthesis via Cascaded Diffusion Models}.
\newblock
\showeprint{2311.04145}


\bibitem[Zhang et~al\mbox{.}(2024)]%
        {zhang2024toratrajectoryorienteddiffusiontransformer}
\bibfield{author}{\bibinfo{person}{Zhenghao Zhang}, \bibinfo{person}{Junchao Liao}, \bibinfo{person}{Menghao Li}, \bibinfo{person}{Zuozhuo Dai}, \bibinfo{person}{Bingxue Qiu}, \bibinfo{person}{Siyu Zhu}, \bibinfo{person}{Long Qin}, {and} \bibinfo{person}{Weizhi Wang}.} \bibinfo{year}{2024}\natexlab{}.
\newblock \bibinfo{title}{Tora: Trajectory-oriented Diffusion Transformer for Video Generation}.
\newblock
\showeprint[arxiv]{2407.21705}~[cs.CV]


\bibitem[Zhao and Zhang(2022)]%
        {zhao2022thin}
\bibfield{author}{\bibinfo{person}{Jian Zhao} {and} \bibinfo{person}{Hui Zhang}.} \bibinfo{year}{2022}\natexlab{}.
\newblock \showarticletitle{Thin-Plate Spline Motion Model for Image Animation}. In \bibinfo{booktitle}{\emph{IEEE Conf. Comput. Vis. Pattern Recog.}} \bibinfo{publisher}{{IEEE}}, \bibinfo{pages}{3647--3656}.
\newblock
\href{https://doi.org/10.1109/CVPR52688.2022.00364}{doi:\nolinkurl{10.1109/CVPR52688.2022.00364}}


\bibitem[Zhou et~al\mbox{.}(2024)]%
        {zhou2024storydiffusion}
\bibfield{author}{\bibinfo{person}{Yupeng Zhou}, \bibinfo{person}{Daquan Zhou}, \bibinfo{person}{Ming{-}Ming Cheng}, \bibinfo{person}{Jiashi Feng}, {and} \bibinfo{person}{Qibin Hou}.} \bibinfo{year}{2024}\natexlab{}.
\newblock \showarticletitle{{StoryDiffusion}: Consistent Self-Attention for Long-Range Image and Video Generation}. In \bibinfo{booktitle}{\emph{Adv. Neural Inform. Process. Syst.}}, \bibfield{editor}{\bibinfo{person}{Amir Globersons}, \bibinfo{person}{Lester Mackey}, \bibinfo{person}{Danielle Belgrave}, \bibinfo{person}{Angela Fan}, \bibinfo{person}{Ulrich Paquet}, \bibinfo{person}{Jakub~M. Tomczak}, {and} \bibinfo{person}{Cheng Zhang}} (Eds.), Vol.~\bibinfo{volume}{37}. \bibinfo{pages}{110315--110340}.
\newblock
\urldef\tempurl%
\url{http://papers.nips.cc/paper\_files/paper/2024/hash/c7138635035501eb71b0adf6ddc319d6-Abstract-Conference.html}
\showURL{%
\tempurl}


\end{thebibliography}


\end{document}